\documentclass{article}

\PassOptionsToPackage{numbers, compress}{natbib}

\usepackage[final]{neurips_data_2022}

\usepackage[utf8]{inputenc} %
\usepackage[T1]{fontenc}    %
\usepackage{hyperref}       %
\hypersetup{
    colorlinks = true,
    linkcolor ={blue},
    anchorcolor ={black},
    citecolor = {black},
    filecolor = {blue},
    breaklinks = true,
    urlcolor = {blue}}
\usepackage{url}            %
\usepackage{booktabs}       %
\usepackage{amsfonts}       %
\usepackage{nicefrac}       %
\usepackage{microtype}      %
\usepackage[table]{xcolor} 
\usepackage{graphicx}
\usepackage{float}
\usepackage{anyfontsize}
\usepackage{csquotes}

\newcommand*\samethanks[1][\value{footnote}]{\footnotemark[#1]}
\usepackage{longtable}
\newcolumntype{L}[1]{>{\raggedright\let\newline\\\arraybackslash\hspace{0pt}}m{#1}}

\title{
{\fontsize{17}{20}\selectfont Pile of Law: Learning Responsible Data Filtering from the Law and a 256GB Open-Source Legal Dataset}
}

\author{%
Peter Henderson\thanks{Equal contribution.}, Mark S. Krass\samethanks[1], Lucia Zheng, Neel Guha\\
\textbf{Christopher D. Manning, Dan Jurafsky, Daniel E. Ho} \\
Stanford University
}

\newcommand{\pilegb}{$\sim$256GB}

\begin{document}

\maketitle

\begin{abstract}
    One concern with the rise of large language models lies with their potential for significant harm, particularly from pretraining on biased, obscene, copyrighted, and private information. Emerging ethical approaches have attempted to filter pretraining material, but such approaches have been ad hoc and failed to take context into account. We offer an approach to filtering grounded in law, which has directly addressed the tradeoffs in filtering material. First, we gather and make available the Pile of Law, a \pilegb{} (and growing) dataset of open-source English-language legal and administrative data, covering court opinions, contracts, administrative rules, and legislative records. Pretraining on the Pile of Law may help with legal tasks that have the promise to improve access to justice.
    Second, we distill the legal norms that governments have developed to constrain the inclusion of toxic or private content into actionable lessons for researchers and discuss how our dataset reflects these norms. 
    Third, we show how the Pile of Law offers researchers the opportunity to learn such filtering rules directly from the data, providing an exciting new research direction in model-based processing.

    \textit{Warning}: this paper contains quotations that may be offensive or upsetting.
\end{abstract}

\section{Introduction}

The presence of private and toxic content in the most popular corpora for pretraining large language models is a well-known problem~\citep{bender2021dangers,jernite2022data}.
But what to do about it is largely a matter of researcher discretion. 
Some teams implement extensive processes for filtering content deemed toxic or private; others train on data in virtually unmodified form.
Resolving all of the difficulties and nuances of content filtering can be challenging, potentially explaining why content filtering has been so uneven.

It is practically difficult to perform reliable and transparent filtering at scale. That is partially because undesirable content is deeply contextual. For example, whether the inclusion of a racial epithet in a dataset is toxic may depend on factors such as the identity of the speaker and
the expectations of the readers \citep{ford2021racial, sperber1981irony}. 
Likewise, the existence of privacy violations may depend in part on the extent to which a speaker expected a fact to be widely shared at the time it was expressed \citep[6-7]{brown2022mean}.
And privacy expectations may vary widely across countries \citep{bellman2004international}. 

Any filtering process involves complex trade-offs. 
Filtering for toxicity may have unexpected effects on representation in datasets or the bias of downstream outputs \citep{dodge2021documenting, gururangan2022whose, bagdasaryan2019differential}. 
And filtering too widely for privacy may harm important downstream applications, as when the Census Bureau's adoption of differential privacy led to errors in redistricting U.S. Congressional districts \citep{ruggles2019differential}.

Yet researchers are not the first to balance the merits of open-source transparency with potential harms: legal and administrative actors have expended significant resources and process in developing standards to strike this exact balance. 
In this work, we suggest that researchers can look to these long-developed (and debated) standards to help ground content filtering mechanisms for large language model training.

This paper makes three contributions. First, we curate and open-source a \pilegb{} (and growing) dataset of legal and administrative data, which we call Pile of Law, which can be used for assessing norms on data sanitization across legal and administrative settings. 
This dataset can be an exploratory tool for evaluating different mechanisms for ``doing the data work''~\citep{sambasivan2021everyone}. And we note that pretraining on the Pile of Law may help with challenging legal tasks that have the potential to improve access to justice~\citep{bommasani2021foundation}. Second, we catalog how government has, though extensive legislation, regulation, and litigation, developed standards for handling the trade-offs between privacy and offensive content on the one hand and transparency, access, and completeness on the other.
We suggest actionable insights for researchers based on these legal and administrative norms.\footnote{Note: while we discuss a number of privacy and toxicity standards, due to the expertise of the authors and the availability of data, this work focuses on the U.S. legal system. We address this and other limitations in Appendix~\ref{app:limitations}.} Third, we demonstrate how implicit sanitization rules can be learned from the Pile of Law, providing a path forward for researchers to develop more nuanced filtering mechanisms. We also demonstrate shortcomings in alignment for current sanitization techniques, providing exciting new directions for research.

\section{Pile of Law}

We curate a \pilegb{} (and growing) dataset of legal and administrative text.\footnote{\href{Available at https://huggingface.co/datasets/pile-of-law/pile-of-law}{https://huggingface.co/datasets/pile-of-law/pile-of-law}.} The utility of this data is twofold: (1) to aggregate legal and administrative data sources that demonstrate different norms and legal standards for data filtering; (2) to collect a dataset that can be used in the future for pretraining legal-domain language models, a key direction in access-to-justice initiatives~\citep{bommasani2021foundation}. A number of prior works have pretrained smaller models on smaller subsets of legal data, including private data that is subject to restrictive licenses~\citep{chalkidis2021lexglue,zheng2021does}. 
None of these have conducted an analysis of the legal data itself---and none have curated an open-source, legal-focused pre-training dataset at this scale.

Through extensive efforts, we compile data from 35 data sources, including legal analyses, court opinions and filings, government agency publications, contracts, statutes, regulations, casebooks, and more. Others have aggregated smaller subsets of legal data, such the EuroParl datasets which gather European Parliamentary debates~\citep{koehn2005europarl,iranzo2020europarl}. We have included some of these as subsets of Pile of Law when relevant and plan to continue adding material to the Pile of Law over time, further increasing its utility to the community. 

We characterize the dataset in detail in Appendix~\ref{app:pile_of_law}. All of the content is already entirely public and mostly available under permissive licenses, but has not previously been compiled at scale for research purposes.\footnote{See Appendix~\ref{app:copyright} for a discussion of copyright and licensing in the dataset.} Each of these data sources carries with it an implicit filtering mechanism formed under relevant legal standards of privacy and toxicity, which we discuss throughout subsequent sections and in the Appendix.
While the underlying data in Pile of Law is already public record and has implicit filters, we recognize that it may contain sensitive material that has escaped administrative scrutiny. We discuss the ethics of our work and our proposed mechanisms for content removal in Appendix~\ref{app:ethics}.

This dataset has obvious utility for pretraining legal-domain foundation models, particularly since, unlike other pretraining data, all material is under open licenses. Though not central to our work, we demonstrate this potential by training an initial BERT-large equivalent model on Pile of Law, yielding comparable results to highly context specific (but smaller) models (see Appendix~\ref{app:models} for full results). 
Recent research has shown in legal contexts that pretraining smaller models on highly in-domain data may be better than large models on big data~\citep{zheng2021does,chalkidis2020legal}.
But in theory, there should be generalizable knowledge and skills that can be learned by training across more diverse sources of data.
A well-crafted pretraining procedure that instills analogical reasoning abilities, for example, should transfer across domains.
Our dataset is large and diverse enough (covering distinct areas of law like criminal law, contracts, and administrative law) to test this hypothesis in the legal domain, where our initial models can form a baseline.

\section{What Can the Law Teach Us About Content Filtering?}
\label{sec:law}

When releasing internal documents concerning individuals, courts and governments have long struggled to balance transparency against the inclusion of private or offensive content.
Model creators now face a similar struggle: what content to filter before pretraining a large language model on the data.  
In this section we survey how governments and courts have handled such content filtering and briefly discuss how Pile of Law implicitly encodes these privacy and toxicity rules.
Based on these rules, we provide actionable lessons for researchers training large language models across fields, so that they can adapt similar rules as minimum standards for dataset sanitization. 
To be clear, we do not take the position that legal rules are optimal nor monolithic. But in many cases they result from a deliberative process that includes judges, legislators, and policymakers in contexts open to public scrutiny, so we think that the machine learning community can at minimum learn from these laws, rules, and norms to improve current ad hoc practice. In short, there is no need to reinvent law.  

\begin{table}[t]
\setlength{\tabcolsep}{4pt} 
\renewcommand{\arraystretch}{1.05}
\small
\caption{Filters Applied in Major Pre-Training Papers}
\begin{tabular}{>{\raggedright}p{0.125\linewidth}p{0.05\linewidth}p{0.15\linewidth}p{0.15\linewidth}p{0.4\linewidth}}
\toprule
& PSI & Deduplication &Toxic Content & Quality \\
 \midrule
\textbf{CCNet} \cite{wenzek2019ccnet} & No & MinHash (pages) & No & No  \\
\textbf{C4} \cite{raffel2019exploring} & No & Unknown (3-sentence spans) & Word list & Minimum word counts, presence of curly brackets, `lorem ipsum', etc.\\
\textbf{GPT-3} \cite{brown2020language} & No & MinHash (pages) & No & Train classifier to distinguish CC from curated high-quality examples  \\
\textbf{Gopher} \cite{rae2021gopher} & No & MinHash (pages) & Google SafeSearch & Min./max. word counts, word-to-symbol ratio, share ellipses, excessive repetition; require stop words \\
\textbf{The Pile} \cite{gao2020pile} & No & MinHash (pages) & Ad-hoc source deletion & Train classifier to distinguish CC from curated high-quality examples\\
\bottomrule
\end{tabular}
\label{tab:lit_review}
\end{table}

\subsection{Privacy} 

Despite the growing focus on privacy in NLP \cite{brown2022mean}, Table~\ref{tab:lit_review} shows that many major pre-training papers do not explicitly filter potentially sensitive information (PSI).\footnote{We define PSI to mean information that could violate a person's privacy interests. This could include personally identifiable information, including a person's name, date of birth, or identification number. Under this definition, a document can contain some PSI (e.g. a name or the facts of a case) while excluding other PSI (e.g. date of birth). But some information that is personally identifiable is not PSI; for example, the name and office contact information of an attorney filing a court brief is identifying but not sensitive.} For example, \cite{gao2020pile} excludes sources due to concerns over explicit or racist content, but does not assess the prevalence of PSI, despite including web-based sources (e.g. OpenWebText) in which users may have an expectation of anonymity.  
Instead, pre-training papers have focused their attention on alternatives to filtering, like deduplication \cite{kandpal2022dedupe}, federated learning \cite{shokri2015privacy,hard2018federated}, differential privacy \citep{lyu2020differentially,feyisetan2020challenges}, and other approaches~\citep{coavoux2018repn,li2018robust,krishna2021adept,yue2021differential,mcmahan2018general}.
But a number of recent papers have demonstrated that large generative models output memorized content \citep{carlini2021extracting,songraghunathan2020leakage,coavoux2018repn,carlini2019secret,lee2022language} even with deduplication~\citep{carlini2022quantifying}. 
Given that many models are trained without privacy mechanisms, filtering is critical to protecting individuals, which is perhaps why research involving health data still emphasizes that approach \citep{norgeot2020protected, adams2019anonymate}. But choosing what to filter is challenging; below, we discuss how governments and courts make such decisions.

\paragraph{How have governments balanced privacy against competing values?} First, we examine how several jurisdictions handle privacy filtering. Table~\ref{tab:comparative_privacy} provides a brief summary.\footnote{See Appendix~\ref{app:law_comp} for a complete version of the table, including citations.} 

\begin{table}
\renewcommand{\arraystretch}{1.2}
\caption{Availability of Identifying Information Across Administrative Settings}
\begin{tabular}{>{\raggedright}p{0.15\linewidth}p{0.25\linewidth}p{0.25\linewidth}p{0.25\linewidth}}
\toprule
\textbf{Jurisdiction} & \textbf{Civil Cases} & \textbf{Criminal Cases} & \textbf{Juvenile Data} \\
 \midrule
U.S. Federal Courts & All case details public unless sealed, except DOBs, ID/account \#s. & Def. names public; DOBs, ID/account \#s, addresses redacted. & Criminal records confidential. Names redacted from civil cases.\\
U.S. Admin. Agencies & Most PII omitted from public records. & - & No statute; more protection in practice.\\
German Courts & Judgments omit all identifying information.  &  Confidential 3-5 years after sentence completed. & No public access to criminal records.\\
Chinese Courts & Names/case details public except in certain classes of cases.  &  Names/case details are public as of 2016.  & Juvenile criminal records are categorically exempt from disclosure. \\
Canadian Courts & Presumption of openness, except specific details and rare sealed cases. &  Public; may be sealed after a period of good behavior.  & Youth criminal records are always confidential. \\
\bottomrule
\end{tabular}
\label{tab:comparative_privacy}
\end{table}

\emph{Baseline Redactions.} Across the jurisdictions we examine, there is a baseline level of filtering. Virtually every jurisdiction in Table~\ref{tab:comparative_privacy} protects the identities of minors. At minimum, juveniles must be protected by pseudonyms in public judgments, and outside of some U.S. states, juvenile criminal records are not public. No jurisdiction normally permits the publication of financial account numbers, dates of birth, or identity numbers like social security numbers.\footnote{The United States' Federal Rule of Civil Procedure 5.2 lays out exceptions when these facts are contained in judicial records properly before a federal court and for civil asset forfeiture cases.} All of these are bright line rules directly applicable to text corpora.

\emph{Value-system contexts.} There are also significant points of disagreement corresponding to the role of privacy in different value systems. U.S., Chinese and Canadian courts denote the names of litigants in ordinary civil cases, prioritizing public access and transparency; German courts do not. Likewise, U.S. federal courts virtually never remove criminal cases from the public record \citep[p. 1233]{shlosberg2011expungement}, a rule also emerging in China \citep{liebman2020mass}. Canada allows most criminal records to be expunged after a period of good behavior. And in Germany, virtually all criminal records are automatically sealed after a set time, and courts have even imposed fines for publicizing a person's criminal history after  expungement \citep{burchardt2020backlash}.

\emph{Contextualized privacy.} Digging further into these rules highlights how court privacy rules account for context. In the U.S. and Canada, the public disclosure of litigants' potentially sensitive information (PSI) can be avoided by persuading a court that extenuating circumstances apply \citep{volokh2021pseud, sccSherman}. To name one example, courts generally permit pseudonyms when parties allege that they have suffered a sexual assault \citep[p. 57]{volokh2021pseud}. The chance to \textit{seal} a case, or to file pseudonymously, suggests that even the most open judicial regime allows for censoring in exceptional cases---although the sealing and pseudonymous filing standards suffer from inconsistency and misuse \citep{sealingreuters}. %

Likewise, administrative agencies often employ context-aware heuristics when deciding whether to include PSI in public decisions.
Although administrative courts are not generally bound by stringent privacy rules like HIPAA \citep{ssaHipaa,vaHipaaMemo}, the Department of Justice exempts immigration applications from public scrutiny due to privacy concerns \citep{morawetz2017better}; the same is true of Social Security Disability applications. Cases involving veterans' benefits are released pseudonymously. 

\emph{Public availability is not a limit.} In many cases, the rules for sanitizing PSI and sealing cases do not depend on whether information is already public. For example, the ban on publicly filing documents revealing dates of birth in U.S. federal courts does not depend on whether a litigant's birth date is otherwise public \citep{eacconfidential}. In cases where a court does take into account the public availability of information (e.g., sealing standards \citep{sccSherman, ca2anon}), contextual countervailing factors can justify keeping a case sealed. 

\paragraph{Implications for Pile of Law.} All of the above privacy norms mean that each subset of Pile of Law is already filtered for privacy based on legal norms in that jurisdiction. Further filtering could seek to align the whole dataset with the norms in one of the subsets prior to pretraining. Appendix~\ref{app:pile_of_law} summarizes the filtering norms present in each subset of the data.

\paragraph{Lessons for researchers.} First, the law provides a number of useful heuristics that researchers could deploy to sanitize data.
Detecting and redacting juvenile names, dates of birth, and account and identity numbers is virtually always appropriate across countries. 
Legal protections for already-public information show why sanitization may be necessary even for text collected from public-facing web pages. 
Second, the U.S. system appears to lean most heavily toward transparency. 
We suggest that researchers can use the U.S. court rules as a floor. Such privacy filtering rules would already go beyond much of current modeling practice. 
Third, in addition to consensus heuristics, researchers should make contextualized decisions about privacy harms. 
While this may seem difficult, Section~\ref{sec:exps} demonstrates how to leverage Pile of Law to learn contextualized standards to mimic legal privacy redaction mechanisms; alternatively, allowing individuals whose information appears in the training corpus to request removal may serve as another stopgap.
Last, the U.S. legal rules do not extend as far as some researchers suggest is necessary. For example, \citep{aura2006scanning} suggests that \textit{all} names must be redacted to preserve privacy. This would reflect greater privacy protection than is typically afforded by U.S. law, which prioritizes public openness and transparency about court proceedings, but would be in line with German rules.  %
These pose important value tradeoffs, and we suggest that researchers look as a starting point to the jurisdiction that aligns with such value tradeoffs for filtering  other potentially sensitive content.

\subsection{Toxicity}

\paragraph{How is toxic speech defined in research?} The category of `toxic speech' is defined in multiple ways \citep{grondahl2018all,vanakenChallenges,ashraf2021abusive}. Some papers define toxicity as ``\textit{disrespectful} comments, including $\dots$ identity \textit{attacks}, profanity and \textit{threats},'' thus emphasizing the idea of intentional insult \citep{dixon2018measuring, yin2021towards, zhang2022opt}.
A broader definition would incorporate \textit{implicit} toxicity, as when a speaker ``subtly'' or ``unconsciously expresses a prejudiced attitude'' \citep{breitfeller-etal-2019-finding, lees-etal-2021-capturing}. \citep{breitfeller-etal-2019-finding} cites the example of the question ``But where are you from, originally?'' 
Others would take a still broader view, suggesting that any \textit{profanity} is toxic (in addition to hate speech and derogatory content) irrespective of speaker intent~\citep{nobata2016abusive}. One implication of these divergent choices concerns \textit{mentions} of toxic language, where a speaker refers to something said by another \citep{sperber1981irony}. For example, if a judge writes that ``Plaintiff claims that her supervisor called her `\_\_\_''' (where \_\_\_ is a profane epithet), an intent-based standard typically would not deem the use of \_\_\_ `toxic,' while an approach targeting profanity typically would. 

\paragraph{How have governments regulated toxic content?} Scholars have documented the role of the law in institutional racism and other forms of oppression, and legal materials from prior eras use words that would by modern standards be considered epithets \citep{cho2008post, berry2011pig}. %
Today, the legal profession in most Anglophone countries strongly polices overt discriminatory epithets \citep{dicosmo2018racism}. Overtly biased speech is prohibited for judges and lawyers in the U.S., Canada, and the U.K. by professional rules \citep{modelcodejudicial, canadianjudicial, bsbHandbook, lsoHandbook}; similar norms have been put forward by the U.N.~\citep{judicial2007commentary}. Judges and lawyers in all countries are routinely sanctioned when they use racist epithets, and most incidents occur verbally or off the bench \citep{dicosmo2018racism, johnson2011racial, judicialconduct2015}. 

Unlike overt, indecorous epithets, legal norms permit the use of speech affected by implicit bias; the incidence of such speech is well-documented ~\citep{prasad2017implicit,rice2019racial, miller2020judicial, benedet2019judicial}. Indeed, it is sometimes encoded in the laws judges and administrative agencies enforce \citep{cahn1991looseness}. Furthermore, some lawyers may see themselves as professionally \textit{obligated} to deploy stereotypes when doing so may assist their clients (e.g. immigration~\citep{morgan2006not}, defendants in sexual assault cases~\citep{craig2013ethical}).

\paragraph{Implications for Pile of Law.} 
The adversarial legal system in many Anglophone countries creates incentives for lawyers to complain about overt racism in written materials, which would violate unambiguous professional rules. Thus, the appearance of epithets in our data is more likely to be confined to quotations, mentions, or to historical legal materials. However, text in our corpus may be toxic according to other definitions; for example, we are unable to quantify the prevalence of implicit biases or offensive stereotyping. Explicit racial, sexual, or offensive terms do appear in modern legal text, but most often in the form of a quotation than direct use. For instance, many cases revolve around evidence documenting racial or gender discrimination, and judges commonly spell out profane or explicit words from the evidentiary record \citep{kennedy2021new, garfield2012swear}. 
Finally, elected officials in our legislative transcripts are not bound by the same professional norms as attorneys.
Additionally, an interesting future examination may note differences between civil law and common law systems, examining rates of offensive content between the different legal systems and norms.
We provide a per-subset examination of filtering norms in Appendix~\ref{app:pile_of_law}.

\paragraph{Lessons for researchers.} First, as is true of privacy, the toxicity norms prevalent in many legal systems offer a lower-bound for researchers. Researchers seeking to mimic the standards that apply in courts should sanitize intentional uses of derogatory terms from pretraining data. That said, current filters are not precise enough to handle this standard. Under the rules applicable to lawyers, filters based on simple word lists would be over-inclusive because they would capture \textit{references} to offensive language that may be non-toxic in context.
Second, the rules applied in courts suggest that generative models should portray toxic behavior explicitly in some contexts, either to serve the values of `accuracy and precision' or to persuade readers \citep[p. 7]{kennedy2021new}; but as \citep{ford2021racial} argues, this view is contested.

Third, in language model pretraining, there may be reason to exceed minimum judicial standards depending on the length of content needed to contextualize references to offensive speech. Accessible language models like Roberta~\citep{liu2019roberta} have a maximum context window of 512 tokens. If a reference to offensive content spans the majority of these tokens, the model will simply uptake the offensive content as if it were being trained for \emph{direct use.} As model contexts grow, it may become more reasonable for researchers to adopt judicial norms.

\section{What Can We Learn from Legal Text?}
\label{sec:exps}

As Section~\ref{sec:law} shows, even jurisdictions that impose a strong presumption of transparency on legal documents often allow for contextual decisions that weigh this presumption against the potential harms caused by the inclusion of PSI on the public record. 
Reducing these rules to tools that can be deployed for filtering may be challenging.
But Pile of Law encodes these contextual decisions already, providing a rich opportunity to learn context-aware norms directly.
This section demonstrates the promise of Pile of Law for operationalizing legal norms. While not comprehensive, the experiments below demonstrate a path forward for replicating the content-filtering mechanisms of courts and governments by leveraging variation in Pile of Law. 
In particular, we show that: (1) Pile of Law reflects variation in privacy norms that can be leveraged to learn contextual privacy rules, such as when to redact names in potentially harmful situations; (2) Pile of Law reflects variation in toxicity norms over time and across contexts, toxicity filters fall short of handling these nuances, and researchers can learn much from building toxicity filters that can handle nuances in Pile of Law's text.

\subsection{Learning Contextual Privacy Rules} 

\begin{figure}
    \centering
    \includegraphics[width=\textwidth]{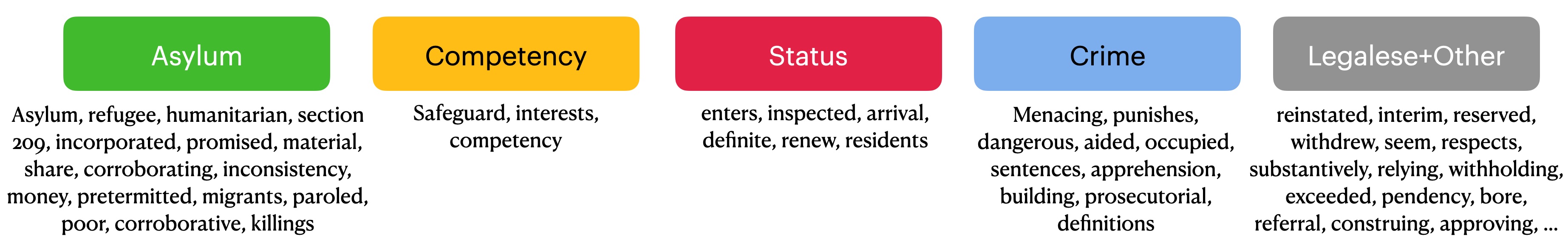}
    \includegraphics[width=.4\textwidth]{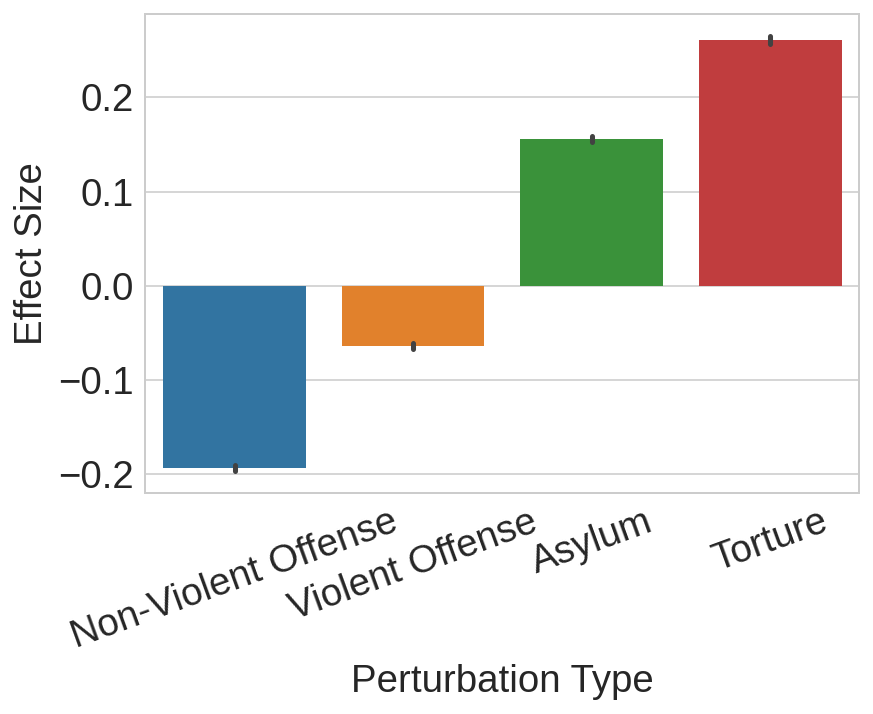}
    \hspace{2em}
    \includegraphics[width=.4\textwidth]{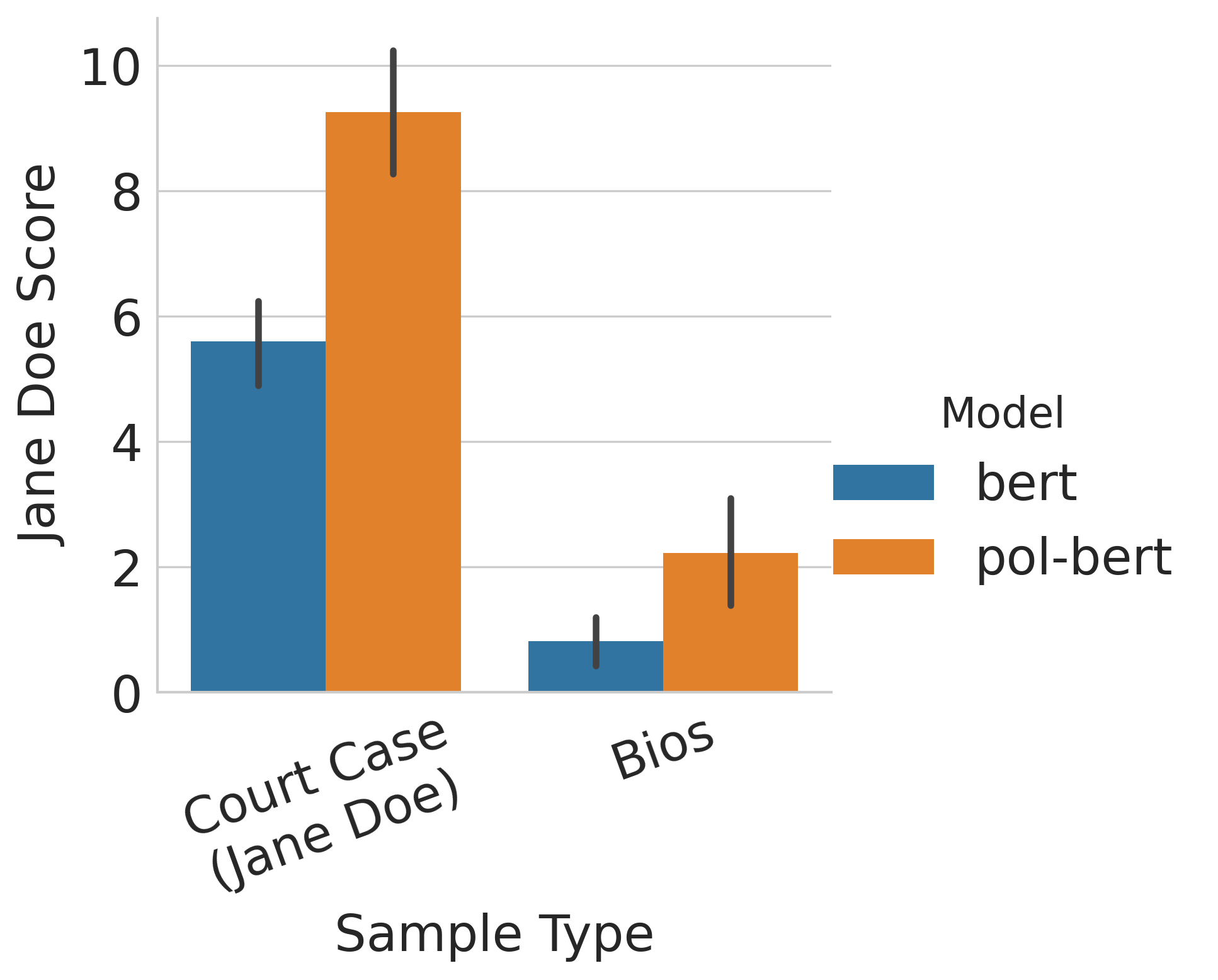}
    \caption{(Top) A causal lexicon learned for the EOIR privacy task, manually sorted by topic with contextual information. Extended version and information in Appendix~\ref{app:eoir}. (Bottom Left) A distill-bert model is more likely to predict pseudonymity for bios with an asylum or torture perturbation (effect size is difference in pseudonymity likelihood from normal bio and bio with added perturbation). (Bottom Right) Jane Doe Score is the difference in MLM score between a version of the sentence using Jane Doe and a random name. The sample sources are paragraphs using pseudonyms and Bios~\citep{de2019bias} (no pseudonyms).}
    \label{fig:lexicon}
\end{figure}

\textbf{Case Study 1: Pseudonymity in Immigration Court.} The Board of Immigration Appeals (BIA) evaluates petitions appealing immigration decisions and sometimes publishes precedential decisions that affect future cases. Some cases include applicants' full names, while others replace them with pseudonymous initials. We demonstrate how subsets of the data can be used to learn the value judgements made in making this pseudonymity decision. We split cases into paragraphs and mask terms used to refer to the applicant. We train a distill-BERT base model~\citep{sanh2019distilbert} to predict whether the paragraph should use pseudonymity or not. This model achieves $\sim$80\% F1 on the validation set. We then examine what types of content are more likely to trigger a pseudonymity recommendation by conducting a perturbation analysis. We use the Bias in Bios dataset~\citep{de2019bias}, censored for names and pronouns. We prepend an additional sentence to each biography that indicates whether the person: (1) is seeking asylum or is a refugee; (2) experienced torture; (3) committed a non-violent criminal offense; or (4) committed a violent criminal offense. Figure~\ref{fig:lexicon} shows that asylum and torture sentences were more likely to trigger pseudonymity while criminal offenses were less likely. This aligns with federal regulations that prevent disclosure of information related to asylum or the Convention Against Torture (8 CFR \S~208.6(a)). By contrast, federal regulations allow information disclosure when a criminal proceeding is involved (8 CFR \S~208.6(d)(1)(ii)), though no regulation addresses criminal history.

Next, we fit a causal lexicon using the deep residualization method (and associated library) from \citet{pryzant2018deconfounded}. We control for the year that a case was published since we found that some aspects of privacy standards have shifted year-to-year, which provides a unique opportunity to learn evolving standards of privacy. We select the top 100 most indicative terms for pseudonymity and remove those where the term only showed up in one case. Then we manually examine contexts and cluster terms into categories. We found that terms most likely to be associated with pseudonymity could be largely clustered into: asylum, mental competency (a legal term used to refer to one's ability to stand trial), immigration status, and indications of a criminal proceeding. We also find that many terms associated with general legal language were included, suggested some remaining confounding and the need for more research into text-based causal attribution. These causal lexicons are seen in Figure~\ref{fig:lexicon}.

\textbf{Case Study 2: Pseudonyms in Civil Litigation.} Next, we look to a ``zero-shot'' version of the experiment above in a broader setting. As noted in Section~\ref{sec:law}, litigants in U.S. courts can ask to use pseudonyms like ``Jane Doe'' in court documents, for example in harassment suits. To assess these requests, courts consider contextual factors like ``sensitive and highly personal'' subject matter, minors, or other extenuating circumstances~\citep{volokh2021pseud}. We collect $\sim$ 500 paragraphs where a pseudonym (``Jane Doe'' or ``Jane Roe'') is used from the validation part of the Court Listener Opinions data. %
For each sentence, we create 100 alternative sentences that replace ``Jane Doe'' with a name sampled using 1990 Census probabilities (using \href{https://github.com/treyhunner/names}{\textsc{names}}).
We then compare whether each model is more likely to guess ``Jane Doe'' using MLM Score~\citep{salazar2019masked}.
We repeat this process on the Bios dataset~\citep{de2019bias}.
Figure~\ref{fig:lexicon} shows that a model trained on Pile of Law (pol-bert) ranks Jane Doe $\sim$ 3 points higher than a standard bert-large-uncased on true pseudonym cases. This suggests that models pre-trained on Pile of Law are more likely to encode appropriate pseudonymity norms. To be sure, pol-bert is slightly more biased for Jane Doe use overall, as is to be expected, but its performance gains persist even after accounting for this bias.  

\textbf{Case Study 3: Privacy Standards in Medical Cases.} We examine \textit{inter-source variation} between the Board of Veterans Appeals (BVA) and the Department of Labor's Employee's Compensation Appeals Board (DOL).
Leading tools for data sanitization remove personal health information as defined by HIPAA~\citep{cohen2018hipaa}, including dates or the name of a physician \citep{norgeot2020protected}.  We ran \citep{norgeot2020protected} on all decisions by the BVA and DOL since both adjudicate the extent of applicants' disabilities, though they are not bound by HIPAA ~\citep{vaHipaaMemo}. 
Showing the difficulty of applying sanitization tools out of domain, virtually \textit{all} decisions included information flagged as HIPAA-protected: 99\% included dates; 96\% of BVA and 100\% of DOL decisions included medical facility names. But the two agencies also differed. About $26\%$ of DOL cases but just $0.36\%$ of BVA cases included a physician name. Physician fraud is more common in worker's compensation programs like DOL's \citep{randfraud}, but the BVA relies on the testimony of VA physicians. The transparency in DOL opinions reflects the higher public interest in physician accountability. 

\textbf{Lessons for researchers.} These experiments show that the Pile of Law encodes signals about privacy standards that can be learned to produce more nuanced recommendations about filtering. For example, researchers may consider whether to mimic the EOIR standard to remove names in proceedings related to minors, asylum or safety concerns. Or they may wish to learn and apply the more contextual standard that is used in general U.S. litigation, where a complex set of factors is used to justify the exclusion of names from case texts.
Such contextualized filters may help ensure that generative models strike the right balance between accuracy and privacy protection, for example by accurately distinguishing benign releases of names and contact information (e.g., in response to queries about government officials) from harmful ones (sensitive circumstances where harm is plausible). 

\subsection{Calibration and Value-Alignment in Toxicity Filtering}

We also identify three main insights (and challenges) from using toxicity filters on Pile of Law, setting the ground for new research using the dataset: (1) toxicity filters often disagree, creating potential issues for automated filtering; (2) toxicity filters may be value-misaligned when it comes to content that is flagged in Pile of Law; (3) toxicity scores vary highly with the length of the content, making it unclear how to handle long-document filtering.

\begin{figure}
\centering
    \includegraphics[width=0.95\textwidth]{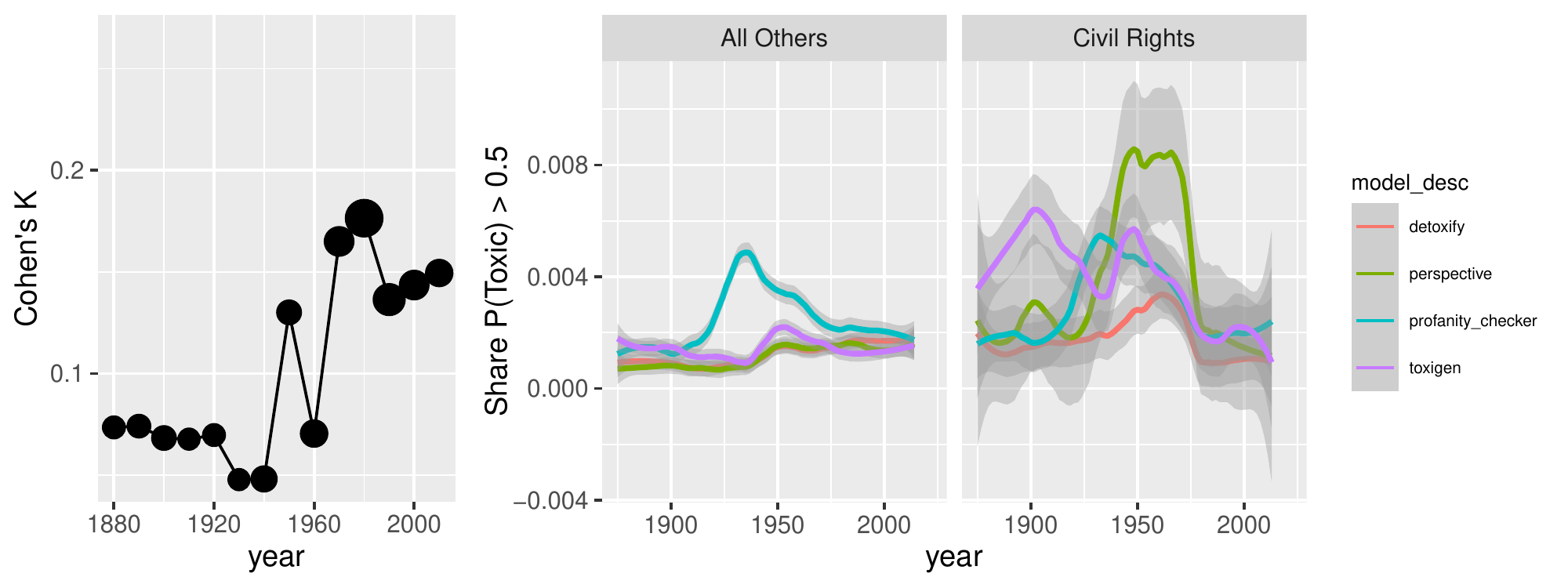}
    \caption{Inter-Model Agreement and Toxicity Over Time. Left: Cohen's $\kappa$, by 10-year bin, calculated for \citep{profanity-check} and \citep{Perspective}, with dot size proportional to number of examples. Right: Share of sentences assigned >50\% probability of being toxic, by model, time, and topic classification \citep{spaeth2013supreme}.}
    \label{fig:toxicity}
\end{figure}

\textbf{Case Study: Supreme Court Decisions.} Leveraging Pile of Law, we show that there are profound nuances to filtering toxic content. First, toxicity filters encode value judgements and divergent definitions of toxicity. Figure~\ref{fig:toxicity} shows Cohen's $\kappa$ between profanity-checker and Perspective over time for sentences in Supreme Court cases (Fig.~\ref{fig:cohens_complete} shows the same for all filters). At the sentence level, the tools' agreement rates are very low, but rise over time, indicating the challenge of handling out-of-domain data far away in time. A vivid example of this challenge is provided in Table~\ref{tab:toxicity_scotus}: \textit{Dred Scott} is the most notoriously racist decision in U.S. history~\citep{greene2011anticanon}, but perhaps due to the archaic language of its holding, \textit{none} of the models is sure that it is toxic. 

\begin{table}[t]
\setlength{\tabcolsep}{4pt} 
    \renewcommand{\arraystretch}{1.05}
    \centering
    \caption{Toxicity Ratings of Quotes From the U.S. Supreme Court, Showing Rating Disagreement} 
    \begin{tabular}{L{0.175\linewidth} p{0.525\linewidth}p{0.05\linewidth}p{0.05\linewidth}p{0.05\linewidth}p{0.05\linewidth}}
    \toprule
    \textbf{Case} & \textbf{Quote} & \textbf{(1)} & \textbf{(2)} & \textbf{(3)} & \textbf{(4)}\\
    \midrule
      Hunter v. Erickson (1969) & ``The majority needs no protection against discrimination.'' & 0.02 & 0.05 & 0.00 & 0.81\\
       Loving v. Virginia (1967) & ``[I]f any white person intermarry with a colored person $\dots$ he shall be guilty of a felony and shall be punished by confinement in the penitentiary $\dots$.'' & 0.52 & 0.54 & 0.60 & 0.94 \\
       Dred Scott~v. Sandford (1857) & ``A free negro of the African race whose ancestors were brought to this country and sold as slaves is not a citizen within the meaning of the Constitution.'' & 0.29 & 0.50 & 0.26 & 0.54 \\
       \bottomrule
       \multicolumn{6}{r}{\footnotesize{\textit{Note:} Model \textbf{(1)} is profanity-check \citep{profanity-check}; \textbf{(2)} is Perspective \citep{Perspective}; \textbf{(3)} is  Detoxify \citep{Detoxify}; and \textbf{(4)} is Toxigen \citep{hartvigsen2022toxigen} .}}
    \end{tabular}
    \label{tab:toxicity_scotus}
\end{table}

But civil rights cases illustrate why the disagreement is about conceptual differences, not just domain drift. Figure~\ref{fig:toxicity} shows that the period between 1950 and 1970 is associated with a large spike in the share of sentences deemed toxic in U.S. civil rights cases. This period was associated with the end of \textit{de jure} segregation in the United States \citep{tushnet1993warren}. Many cases likely \textit{quoted} or \textit{mentioned} racist laws before striking them down. For instance, in Table~\ref{tab:toxicity_scotus}, \textit{Loving} describes a law banning interracial marriage in order to deem it illegal. Quoting this language qualifies as toxic under some but not all definitions, and as Figure~\ref{fig:toxicity} shows, that view is encoded in some but not all filters. Accordingly, the filters disagree as to whether \textit{Loving}'s quote is clearly toxic. Document-level filtering could thus easily delete core civil rights cases like \emph{Hunter} and \emph{Loving}---while  leaving in \emph{Dred Scott}.

Finally, we note that the context window used to filter out sentences appears to dramatically influence ratings. Perspective segments data into sentences and then labels each sentence, which is the approach we take above. We find that by using longer span, we can \emph{systematically decrease} the perceived toxicity of a span, even if it is obviously toxic under any definition. We take the top 5k sentences labeled as toxic by Toxigen. We then take 2 sentences before and after the toxic sentence (clamped to the boundaries of the document). We find that the toxicity score drops between \textbf{55-57\%} (absolute, 95\% CI) just by adding this context. While some of this change might be due to correct re-classification of mentions, we provide qualitative examples in which this is clearly untrue in Appendix Table~\ref{tab:toxigen_context_length}.

\textbf{Lessons for researchers.} The experiments above demonstrate that, while toxicity filtering is important to align with the courts' modern lower bounds banning uses of epithets, it is not clear that existing filters are not consistent and filter out content aligned with different values. Moreover, they can arbitrarily label content as non-toxic in long-document or out-of-distribution settings, which may affect filtering mechanisms. More work is needed to create robust, value-aligned toxicity filters for pretraining and it is unclear if off-the-shelf mechanisms strike the right balance. As we have shown, the Pile of Law provides unique opportunities to develop such methods. 
\section{Conclusion}

In this work we have examined how the law and legal data can inform data filtering practices that are of great importance to responsible large language model training. We provide an extensive legal dataset (the Pile of Law) and illustrate a number of exciting new research directions for future work.

\section*{Acknowledgements}

We thank SambaNova Systems for generously providing compute resources via the SambaNova Systems Dataflow-as-a-Service\textsuperscript{TM} platform and the Stanford Institute for Human-Centered Artificial Intelligence for computing support. We also thank Jieru Hu for helpful discussions and Krithika Iyer for technical assistance. PH is supported by an Open Philanthropy Project AI Fellowship.

\bibliographystyle{plainnat}
\bibliography{refs}

\clearpage
\appendix

\section*{Checklist}

\begin{enumerate}

\item For all authors...
\begin{enumerate}
  \item Do the main claims made in the abstract and introduction accurately reflect the paper's contributions and scope?
    \answerYes{}
  \item Did you describe the limitations of your work?
    \answerYes{}
  \item Did you discuss any potential negative societal impacts of your work?
    \answerYes{Please see our ethics statement.}
  \item Have you read the ethics review guidelines and ensured that your paper conforms to them?
    \answerYes{Please see Appendix A, where we provide a point-by-point discussion of how our paper conforms to the 2022 NeurIPS ethics guidelines.}
\end{enumerate}

\item If you are including theoretical results...
\begin{enumerate}
  \item Did you state the full set of assumptions of all theoretical results?
    \answerNA{}
        \item Did you include complete proofs of all theoretical results?
    \answerNA{}
\end{enumerate}

\item If you ran experiments...
\begin{enumerate}
  \item Did you include the code, data, and instructions needed to reproduce the main experimental results (either in the supplemental material or as a URL)?
    \answerYes{See Appendix~\ref{app:code}.}
  \item Did you specify all the training details (e.g., data splits, hyperparameters, how they were chosen)?
    \answerYes{}
        \item Did you report error bars (e.g., with respect to the random seed after running experiments multiple times)?
    \answerYes{}
        \item Did you include the total amount of compute and the type of resources used (e.g., type of GPUs, internal cluster, or cloud provider)?
    \answerYes{See Appendix~\ref{app:carbon}.}
\end{enumerate}

\item If you are using existing assets (e.g., code, data, models) or curating/releasing new assets...
\begin{enumerate}
  \item If your work uses existing assets, did you cite the creators?
    \answerYes{}
  \item Did you mention the license of the assets?
    \answerYes{}
  \item Did you include any new assets either in the supplemental material or as a URL?
    \answerYes{}
  \item Did you discuss whether and how consent was obtained from people whose data you're using/curating?
    \answerYes{}
  \item Did you discuss whether the data you are using/curating contains personally identifiable information or offensive content?
    \answerYes{Please see Appendix A, where we provide a point-by-point discussion of how our paper conforms to the 2022 NeurIPS ethics guidelines.}
\end{enumerate}

\item If you used crowdsourcing or conducted research with human subjects...
\begin{enumerate}
  \item Did you include the full text of instructions given to participants and screenshots, if applicable?
    \answerNA{}
  \item Did you describe any potential participant risks, with links to Institutional Review Board (IRB) approvals, if applicable?
    \answerNA{}
  \item Did you include the estimated hourly wage paid to participants and the total amount spent on participant compensation?
    \answerNA{}
\end{enumerate}

\end{enumerate}

\section{Ethics Statement}
\label{app:ethics}
While we recognize that the NeurIPS Ethical Guidelines are not an exhaustive list, much of our paper concerns the broader ethical questions surrounding the privacy and toxicity harms of our data. Accordingly, we limit this Appendix to specifically addressing the nine points eplicitly mentioned in the 2022 version of the Guidelines. 

\textbf{1. Does the data contain any personally identifiable information or sensitive personally identifiable information?}

The data contains PII. As we argue in the main text, each document presumptively follows the privacy norms of the jurisdiction where it was written, which necessarily means that some documents include PII that would violate the norms of a different jurisdiction (e.g., the inclusion of names in U.S. criminal cases would violate German privacy rules). 
Nonetheless, as we argue in the paper, a reasonable minimum standard for filtering is deference to privacy rules of a particular jurisdiction that already weighs transparency benefits against privacy harms. 
That is especially true given that our data is largely produced by governmental entities, not merely sanctioned by them.
These rules are developed to balance privacy against transparency in a manner specific to cultural context and we respect these standards. 

That said, while others have suggested different data governance strategies~\citep{jernite2022data}, we follow the approach of CourtListener, which has already grappled with the exact trade-off of compiling legal data~\citep{lissner2010courtlistener}. CourtListener runs a filtering mechanism to remove information like SSNs that may have slipped through the courts' first pass redaction layer. CourtListener also provides a mechanism for stakeholders to take down cases. We also run a filter to validate that no Social Security Numbers (SSNs) were present in the data using Microsoft Presidio~\citep{presidio}.
Like them, we check for high-risk information like SSNs and restrict indexing of the data by search engines to the best of our ability. We also provide instructions for requesting content removal on the dataset website that reflect the CourtListener mechanism.
We also have enabled the HuggingFace community feature to allow requests for dataset changes and removal of content.

We do not redact information further for several reasons. Some of the data subsets, in particular the court subsets, request that reproduction ``exercise due diligence in ensuring the accuracy and currency of the materials reproduced'' (see, e.g., \url{https://www.bccourts.ca/Privacy\%20Statement.html}).
By using existing anonymization filters that might noisily redact factual information or legal citations, we cannot ensure such accuracy and concurrency. As such, we chose to respect the data sources’ decisions on the question of balancing privacy against transparency and to reproduce the content in the most accurate way possible. Furthermore, removing names for common law data would break the important legal references and context if done in a non-contextual manner. For example, a quote like ``In \emph{Brown}, Justice Warren wrote...'' would become ``In [MASK], Justice [MASK] wrote...'', which would not provide any information about the underlying case law nor the facts of the case to which the law should apply.

Since the goal of the paper is to provide mechanisms and data for ``doing the data work,'' we largely leave the data unfiltered beyond the mechanisms described above. Because we may be bound by upstream licenses prohibiting further restrictions on use, we place the data compilation under a CreativeCommons Attribution-NonCommercial-ShareAlike 4.0 International license, but underlying data subsets may be bound by other licenses (see Table~\ref{tab:licensing}).

\textbf{2. Does the data contain information that could be deduced about individuals that they have not consented to share?}

It is possible that the case details included in sealed, pseudonymized, or otherwise sanitized case materials could be used to deduce the identity of litigants. Nonetheless, as we note above and in the main text, the degree of protection afforded to litigants in these cases is precisely the protection to which they are entitled by law. 
Furthermore, litigants have ample notice about the degree of protection afforded to their identities in public cases and mechanisms to challenge decisions on anonymity.
We acknowledge that there may be shortcomings to this process, however. Thus, following the mechanism put forth by CourtListener and others, we provide a mechanism via the HuggingFace community feature to request the removal of content. In this process, we ask that stakeholders file a takedown request with the upstream data source and notify us of its modification in that source. Then we will update our archive with the upstream. This ensures that we comply with upstream licenses and takedown standards. 
In exceptional cases where the upstream source does not take down especially harmful data, we will consider such situations on a case-by-case basis.
We also will prevent search engine indexing of the data to the best of our ability.

\textbf{3. Does the data encode, contain, or potentially exacerbate bias against people of a certain gender, race, sexuality, or who have other protected characteristics?}

As we note in the main text, some of the court opinions we include may rely on precedential cases that relied upon past legal regimes including racial or other discrimination. Further, there is empirical evidence of implicit bias among judges \citep{levinson2017judging}. However, as we argue in the paper, there is reason to believe that explicitly derogatory expressions directed at particular vulnerable groups are absent from recent legal text.

Further, there is a high degree of public interest in the materials included in our dataset, because they reflect the public decisions of governmental institutions and may continue to bind future parties. 
For example, civil rights cases we reference (e.g., \emph{Brown}, \emph{Loving}) are crucial civil rights law that legal models must understand despite (or perhaps because of) their references to unjust legal regimes. 
And since we argue that a large benefit of our data is that it provides a new way to examine and address shortcomings in toxicity filters, we leave content unfiltered.

\textbf{4. Does the paper contain human subject experimentation and whether it has been reviewed and approved by a relevant oversight board?}

No.

\textbf{5. Does the paper rely on data that have been discredited by the creators?}

No. 

\textbf{6. Consent to use or share the data. Explain whether you have asked the data owner’s permission to use or share data and what the outcome was.}

Consent for the curated data is implicit in each subset, in particular as indicated by the license it was released, or the legal standard under which it was made public. For example, parties to lawsuits in U.S. federal court have ample notice that their names and case details will become part of the public record, since that is encoded in the rules we reference above. 

\textbf{7. Domain specific considerations when working with high-risk groups.}

As we note in the text of the paper, while high-risk groups appear in this data, we rely upon the filtering mechanisms afforded for such circumstances by the government publishers of the text. While our work may make data about such groups easier to access, it is not obvious that this will cause harm given the significant positive implications of improved access to legal information.

\textbf{8. Filtering of offensive content. For instance, when collecting a dataset, how are the authors filtering offensive content such as racist language or violent imagery?}

There may be racist or toxic language in the dataset, but as we mention in this work, many of these instances arise when a court is describing the facts of the case as opposed to a direct use of epithets. To be sure, as we note in the main text, historical precedents and cases relying on them are more likely to include toxic material given the legal system's historical role in enforcing discriminatory laws. Further, there is empirical evidence of implicit bias among judges \citep{levinson2017judging}. 
And since we argue that a large benefit of our data is that it provides a new way to examine and address shortcomings in toxicity and privacy filters, we leave content unfiltered so that contextual filters can be learned from the data.

We also note that there may well be violent content particularly in discussion of criminal cases, but for legal language models it is often important to understand the facts of a case and apply the law to these facts. As we argue in this work, one path forward is to learn context-dependent filters based on the court's protection of victims that would retain content relevant to the law while filtering out identifiable information for protected individuals.

\textbf{9. Compliance with GDPR and other data-related regulations. For instance, if the authors collect human-derived data, what is the mechanism to guarantee individuals’ right to be forgotten (removed from the dataset)?}

As noted in the paper, our data is derived from governmental sources that impliedly comply with the privacy rules applicable in their respective jurisdictions. GDPR applies only to persons ``monitoring the behavior of individuals in the EU.'' The European data we collect that concerns `individuals' other than Parliamentarians comes from the European Court of Human Rights which must already have complied with relevant provisions. HIPAA does not apply to any of the data we collect. For example, as we note in the main text, although many of the administrative cases we collect concern medical records, the organizations issuing those decisions are not HIPAA custodians and their decisions are not subject to HIPAA protection \citep{vaHipaaMemo}. Nonetheless, we provide a mechanism to take down content as noted above.

\section{Limitations}

\label{app:limitations}

Due to licensing restrictions, our collation of judicial texts from the Anglophone countries is concentrated on U.S. texts (though it does include a number of international sources, including European Parliament proceedings and the decisions of Canadian courts). For example, BAILII, the online repository of judgments issued by British and Irish courts, forbids ``storing search results or HTML versions of judgments.''\footnote{See https://www.bailii.org/bailii/copyright.html} We encourage future efforts to identify freely-available versions of legal English-language texts in other countries to improve the geographic coverage of the data.

Furthermore, our experiments are meant to be demonstrative and set a research agenda. The learned filters we describe in Section~\ref{sec:exps} will likely require more work to be applied to significantly out of domain data (though we show that they are somewhat robust through a perturbation analysis). We believe this is an exciting new research pathway.

\section{Carbon Impact Statement}
\label{app:carbon}

As suggested by \citet{henderson2020towards}, \citet{lacoste2019quantifying} and others, we report the
energy and carbon impacts of our experiments. While we are unable to calculate precise carbon emissions from hardware counters, we give an estimate of our carbon emissions. 
For the BERT models, we estimate roughly 8 weeks of RDU usage total at full capacity, including debugging, hyperparameter optimization, iteration on experiments on 8 RDUs. 
We are unsure of the exact TDP for these chips, but as rough estimate assume that the TDP of each is similar to an A100 GPU or a TPUv2. Assuming minimal CPU time, this is roughly 4704 kWh of energy used and 1121 kg  CO$_{2eq}$  at the California yearly average carbon intensity of 238.4 g CO$_{2eq}$ / kWh.
Other experiments took relatively smaller amounts of compute using various CPUs and GPUs, mostly at inference time. Fine-tuning of the BERT-Large model on CaseHOLD used 4 A100s for 2 days. Other experiments ran inference on a variety of GPUs/CPUs. To put a conservative estimate on this, we assume roughly 4 weeks of total added compute time at full capacity. Assuming an average TDP of 300W across these machines, this yields 201.6 kWh and 480.6 kg  CO$_{2eq}$.
We emphasize that these are extremely rough estimates (and may be overestimates due to their assumptions) and they take into account all experimentation, including iterations and debugging.
\section{Code and Data Availability}
\label{app:code}
Pile of law is hosted at \href{ https://huggingface.co/datasets/pile-of-law/pile-of-law}{https://huggingface.co/datasets/pile-of-law/pile-of-law}. 
The main pol-bert checkpoint we use in the paper is available at \href{https://huggingface.co/pile-of-law/legalbert-large-1.7M-2/}{https://huggingface.co/pile-of-law/legalbert-large-1.7M-2/} and an additional experimental checkpoint using a different random seed yet otherwise identical settings is at \href{https://huggingface.co/pile-of-law/legalbert-large-1.7M-1/}{https://huggingface.co/pile-of-law/legalbert-large-1.7M-1/}. We will also make intermediate checkpoints (every 50k timesteps) available on request.
The EOIR experiment data and trained model are available at \href{https://huggingface.co/datasets/pile-of-law/eoir\_privacy}{https://huggingface.co/datasets/pile-of-law/eoir\_privacy} and \href{https://huggingface.co/pile-of-law/distilbert-base-uncased-finetuned-eoir\_privacy}{https://huggingface.co/pile-of-law/distilbert-base-uncased-finetuned-eoir\_privacy}, respectively.

Code for experiments and data collection will be available at \href{https://github.com/Breakend/Pileoflaw}{https://github.com/Breakend/Pileoflaw}. We include pre-processing code for the data used for pre-training, but do not provide the pre-training code itself since we directly use the out-of-the-box SambaNova pre-training tool with no changes of our own (other than at pre-processing time).
\section{Pile of Law Data Description}
\label{app:pile_of_law}
We briefly describe the data curated for each subsection, grouped by logical similarity. 

\subsection{Legal Case Opinions and Filings}
\textbf{CourtListener Opinions, CourtListener Docket Entries and Court Filings}. CourtListener provides a large set of U.S. court case opinions across a number of federal and state courts. 
Our dataset includes all \textit{judicial opinions} in the CourtListener opinions dataset. This is similar to the FreeLaw portion of \citep{gao2020pile}. \textit{Judicial opinions} are long-format documents written by judges explaining and justifying a decision about the factual or legal disposition of a case. Judicial opinions are typically written in a more authoritative and less argumentative style.

To supplement the data with argumentative language, we scrape CourtListener's RECAP Docket Entry Documents API from present day until 2018 (due to rate limits we limit to the past few years; future iterations of the dataset may gather more documents). RECAP includes \textit{briefs}, which are a party's arguments to the court, as well as interim judicial opinions, exhibits (i.e., the evidence supporting a party's claim), and other miscellaneous case records. RECAP docket entry documents are any publicly available filings provided on a court docket. Dockets do not generally include evidence that is not referenced by either party. 

\textbf{U.S. Supreme Court Docket Entries and Court Filings}. The U.S. Supreme Court docket system contains information about the status of pending and decided cases that have been filed at the Court. A docket is a log containing the history of each case in the form of brief chronological entries summarizing the court proceedings and the court filings, the underlying documents (pleadings, motions, briefs, etc.) filed with the Court in the proceedings of a case. Though dockets and court filings do not have precedential value, the information contained can sometimes provide additional insight into why the Court issued a particular decision or opinion.

\textbf{U.S. Board of Veterans' Appeals Decisions}. The U.S. Board of Veterans' Appeals (BVA) is an internal administrative agency of the U.S. government that hears appeals from veterans who were denied disability or other benefits by the Veterans' Benefits Administration. Every single case involves a veteran arguing for more benefits, opposed by the government arguing for the status quo. The vast majority of BVA opinions concern either (A) the severity of a veteran's disability or (B) the etiology of the disability and whether it can be traced to their period of service in the U.S. military
\citep{huang2021context}.

\textbf{U.S. Federal Trade Commission Advisory Opinions}. When a business or trade group wishes to engage in a practice that may violate competition or consumer protection laws, it can request an official opinion from the U.S. Federal Trade Commission (FTC), which issues an \textit{advisory opinion}, often written in a similar manner to a judicial opinion.

\textbf{U.S. National Labor Relations Board Decisions}. The U.S. National Labor Relations Board (NLRB) governs the relations between employers and unions, and when one party alleges that there has been a violation of the National Labor Relations Act (e.g., an unfairly held union election), the NLRB is tasked with deciding whether an unfair labor practice has occurred.

\textbf{U.S. Department of Justice Executive Office for Immigration Review \textit{Immigration \& Nationality Decisions}}. When a person disagrees with the decision of an immigration judge, they can appeal to the Board of Immigration Appeals (BIA) and/or directly to the Attorney-General (AG), who may also decide to review BIA decisions independently \citep{hausman2021executive}. We download decisions of the BIA and the AG included in the DOJ's \textit{Immigration \& Nationality Decisions}. 

\textbf{U.S. Tax Court PLR Corpus}.
The U.S. Tax Court resolves disputes between taxpayers and the Internal Revenue Service (IRS). We include the data released by~\citep{taxcorpus}, which includes Tax Court memorandum and summary opinions scraped from the Tax Court website and IRS Private Letter Rulings (PLRs), scraped from the IRS website. PLRs are written statements, issued at the request of a taxpayer, that interprets and applies tax laws to the taxpayer's represented set of facts.

\textbf{U.S. Department of Labor Employees' Compensation Appeals Board Decisions}. When a person disagrees with the decision of the U.S. Department of Labor Office of Workers' Compensation Programs (OWCP), they can appeal to the Employees' Compensation Appeals Board (ECAB). We download decisions of the ECAB from 2006 to April 2022.

\textbf{European Court of Human Rights Opinions}. The European Court of Human Rights (ECHR) hears appeals by individuals or states bound by the European Convention on Human Rights, and issues precedential opinions binding state parties. We include the ECHR corpus, which contains 42 decisions of the ECHR and was created and introduced by~\citep{poudyal2020echr}.

\textbf{Canadian Court Opinions}. The Ontario Court of Appeals (ONCA) and the British Columbia Court of Appeals (BCCA) are intermediate appeals courts that hear all manner of cases originating in their respective jurisdictions. We download the available opinions listed on the ONCA and BCCA websites.

\subsection{Legal Analyses}
\textbf{U.S. Office of Legal Counsel Memos}.
The U.S. Office of Legal Counsel (OLC) advises the U.S. President on the legality of any action the president wishes to take. OLC issues memos that provide in-depth legal analyses on various topics that have confronted presidents in the past. These should not be considered ground truth for the law, but provide good examples of in-depth legal analysis and are similar to judicial opinions in tone.

\textbf{U.S. Department of Justice Inspector General Reports}.
Almost every federal agency in the U.S. federal government has an Office of the Inspector General, which is responsible for independent oversight of the agency. This includes regular audits of the agency's spending, monitoring of active government contractors, and investigations into wasteful or corrupt agency practices, which are compiled and published in public reports. We download the reports of every U.S. IG published online, which were originally scraped by~\citep{mill2014opening}.

\subsection{Laws}
\textbf{U.S. Code of Federal Regulations, U.S. State Codes, U.S. Code, U.S. Federal Rules of Evidence, U.S. Federal Rules of Civil Procedure}.
We scrape the U.S. Code of Federal Regulations, U.S. State Codes, U.S. Code, Federal Rules of Evidence, and U.S. Federal Rules of Civil Procedure. These are statutes and rules that govern much of the U.S. legal system and are important knowledge for machine learning algorithms.

\textbf{U.S. Bills, U.S. Federal Register}.
We scrape proposed U.S Bills and the U.S. Federal Register, which includes proposed and final rules along with other regulatory actions, to imbue understanding of statutory construction (Bills) and regulatory construction (Federal Register). We use the Government Publishing Office's govinfo API and Bulk Data Repository~\citep{govinfo} to scrape proposed U.S. Bills and the U.S. Federal Register. The Bulk Data Repository provides digitized proposed U.S. Bills from the 113th Congress (2013-2014) to the 117th Congress (2021-2022) and proposed Federal Register rules from 2000-2022. We scrape from the beginning of the available time range to October 2021.

\textbf{U.S. Founders Letters}.
The U.S. founders letters describe the birth of the American Republic and its democratic and legal institutions. We scrape $\sim$185,000 documents of correspondences and other writings by U.S founders, George Washington, Benjamin Franklin, John Adams (and family), Thomas Jefferson, Alexander Hamilton, John Jay, and James Madison, from Founders Online~\citep{Founders}.

\textbf{World Constitutions}.
Constitutional construction is an important task for state-building -- as well as for fundamental legal analysis. As such, we add the world's constitutions~\citep{elkins2014constitute} to the data.

\textbf{EUR-Lex}.
EUR-Lex provides access to European Union (EU) legal documents, including treaties, legal acts from EU institutions, preparatory documents, EU case law, EU international agreements, European Free Trade Association (EFTA) documents, national transposition measures, and national case law related to EU law. EUR-Lex provides greater context on legal systems in the European Union and international law. We add EUR-Lex data from \citep{chalkidis-etal-2019-large}, an expanded version of the Eur-Lex dataset released by~\citep{loza2010efficient}.

\subsection{Contracts / Business Documents}
\textbf{Credit Card Agreements, Terms of Service, Edgar Contracts, Atticus Contracts}.
Credit Card Agreements provided by the U.S. Consumer Financial Protection Bureau,\footnote{\url{https://www.consumerfinance.gov/credit-cards/agreements/}}, Terms of Service~\citep{lippi2019claudette,ruggeri2021detecting}, Edgar Contracts~\citep{borchmann-etal-2020-contract}, Atticus Contracts~\citep{hendrycks2021cuad} were all added to the data, which may be useful pretraining data for a large number of contract based tasks that are beginning to employ machine learning methods, such as contract review~\citep{leivaditi2020benchmark, hegel2021law, hendrycks2021cuad}.

\subsection{Conversations}
\textbf{U.S. Congressional Hearings}. A congressional hearing is a meeting or a session of a Senate, House, joint, or special committee of Congress, to obtain information and opinions on proposed legislation, conduct an investigation, evaluate the activities of a government department or the implementation of a federal law, or provide testimony and data about topics of current interest. These hearings provide details on the research and drafting process for proposed legislation. The transcripts of congressional hearings from the 89th Congress (1965-67) to the 117th Congress (2021-2022) were scraped using the Government Publishing Office's govinfo API and Bulk Data Repository~\citep{govinfo}.

\textbf{European Parliament Proceedings Parallel Corpus}.
The European Parliament proceedings capture debate about proposed EU legislation. We include only the English data in the European Parliament Proceedings Parallel Corpus~\citep{koehn2005europarl}, though the original data provides parallel translations in 11 languages.

\textbf{U.S. Supreme Court Oral Argument Transcripts}.
The U.S. Supreme Court holds oral arguments in about 70-80 cases each year. Oral arguments give the Justices an opportunity to ask questions to the attorneys representing the parties to the case and the attorneys to highlight important arguments of the case. We extract oral argument transcript data using raw data made available through \cite{scotus_oral_arguments}, from the 2021-08-14 release.

\textbf{U.N. General Debate Corpus}.
The General Debate is held at the beginning of each session of the United Nations (UN) General Assembly, which has convened annually since 1946. The General Debate is a forum for world leaders and other senior officials representing UN member states to deliver statements that present their government's perspective on the major issues in world politics, analogous to legislative state-of-the-union addresses in domestic politics. We include the General Debate statements from the UN General Debate Corpus (UNGDC)~\citep{baturo2017understanding} in the Pile of Law. The UNGDC contains General Debate statements from 1970 (Session 25) to 2020 (Session 75).

\textbf{Reddit r/legaladvice \& r/legaladviceofftopic}.
Because most legal language is often difficult to understand for the lay person and does not encode clear answers to simple legal questions, we sought to find a dataset that yields a ``plain English'' Q\&A format. We settled on the two subreddits r/legaladvice and r/legaladviceofftopic. Because of the risk of encoding incorrect legal advice, we heavily filtered the data. We filtered out any posts with profanity using profanity-check~\citep{profanity-check}. We also only included posts with at least one answer with a score of over 8 net upvotes. We then restructured the data as:

Title: [Post Title]\\
Question: [Post Content]\\
Topic: [Post Flair]\\
Answer \#[N]: [Top Answers]... 

We used the PushShift API to scrape the entirety of the each subreddit~\citep{baumgartner2020pushshift}.

\subsection{Study Materials}
\textbf{Bar Exam Outlines}.
The bar exam outlines provide key definitions and descriptions of concepts relevant to various subject areas in American law that are tested on the bar examination (e.g., Constitutional Law, Contracts, Criminal Law, etc.). In every U.S. jurisdiction (except in certain cases in Wisconsin), all applicants seeking admission to the bar to practice law in the jurisdiction must pass a bar examination.

\textbf{Open Source Casebooks}.
Legal knowledge is hard to parse from individual cases. Casebooks and textbooks have been created to teach students with edits cases and commentary to focus learning on the most important legal topics. As such we gather $\sim$60 casebooks that were released under a Creative Commons license. The casebooks range across a broad range of topics corresponding to core doctrinal material. All licenses and author credits remain self-contained in the individual documents.

\begin{table}[!htbp]
    \centering
        \caption{Description of the Pile of Law by Data Source}
    \resizebox{\textwidth}{!}{
    \begin{tabular}{p{7cm}ccc}
    \toprule
         Data Source & Data Size & Word Count & Document Count \\
         \midrule
         Court Listener Opinions & 59.29GB/19.76GB & 7.65B/2.55B & 3.39M/1.12M\\
         \hline
         Court Listener Docket Entries and Court Filings & 52.13GB/17.38GB & 5.36B/1.79B & 1.49M/496K \\
         \hline
         U.S. Supreme Docket Entries and Court Filings & 1.51GB/0.50GB & 151.05M/51.73M & 48K/16K \\
         \hline
         U.S. Board of Veterans' Appeals Decisions & 13.21GB/4.40GB & 1.74B/580.98M & 630K/210K  \\
         \hline
         U.S. Federal Trade Commission Advisory Opinions & 1.55MB/0.52MB & 157K/53K & 112/33  \\
         \hline
         U.S. National Labor Relations Board Decisions & 994.83MB/331.61MB & 120.33M/39.20M & 24K/8K \\
         \hline
         U.S. Department of Justice Executive Office for Immigration Review \textit{Immigration \& Nationality Decisions} & 22.89MB/7.63MB & 3.05M/1.01M & 1671/558 \\
         \hline
         U.S. Department of Labor Employees' Compensation Appeals Board & 353.25MB/117.75MB & 48.20M/16.01M & 21K/7K \\
         \hline
         European Court of Human Rights Opinions~\citep{poudyal2020echr} & 111.53MB/37.18MB & 16.71M/3.47M & 7K/1K  \\
         \hline
         Canadian Court Opinions (ON, BC) & 182.09MB/60.70MB & 23.45M/7.66M & 8K/3K \\
         \hline
         U.S. Office of Legal Counsel Memos & 36.98MB/12.33MB & 4.36M/1.31M & 1038/346 \\
         \hline
         U.S. Office of Inspector General Reports & 1.90GB/0.63GB & 167.71M/54.18M & 29K/10K \\
         \hline
         U.S. Code of Federal Regulations & 670.87MB/223.62MB & 79.06M/25.41M & 182/61 \\
         \hline
         U.S. Supreme Court Oral Argument Transcripts & 1.51GB/0.50GB & 151.05M/51.73M & 47K/16K \\
         \hline
         U.S. State Codes & 6.77GB/2.26GB & 829.62M/441.38M & 157/60 \\
         \hline
         U.S. Code & 268.40MB/89.47MB & 30.54M/18.20M & 43/15 \\
         \hline
         U.S. Federal Rules of Evidence & 670KB/223KB & 77K/36K & 51/17  \\
         \hline
         U.S. Federal Rules of Civil Procedure & 1.59MB/0.53MB & 237K/40K & 69/23 \\
         \hline
         U.S. Bills & 1.27GB/0.42GB & 156.06M/49.4M & 84K/28K  \\
         \hline
         U.S. Federal Register & 159.29MB/53.10MB & 6.61M/53.27M & 4060/1354  \\
         \hline
         U.S. Founders Letters & 419.33MB/139.78MB & 53.27M/17.69M & 138K/46K \\
         \hline
         World Constitutions~\citep{elkins2014constitute} & 24.43MB/8.14MB & 3.43M/1.06M & 139/48 \\
         \hline
         EUR-Lex~\citep{chalkidis-etal-2019-large} & 1.31GB/0.44GB & 191.65M/65.31M & 106K/36K \\
         \hline
         Credit Card Agreements & 70.19MB/23.40MB & 10.73M/3.09M & 2023/615 \\
         \hline
         Terms of Service~\citep{lippi2019claudette,ruggeri2021detecting} & 1.57MB/0.52MB & 213K/62K & 37/13 \\
         \hline
         Edgar Contracts~\citep{borchmann-etal-2020-contract} & 10.76GB/3.59GB & 1.44B/473.50M & 741K/247K \\
         \hline
         Atticus Contracts~\citep{hendrycks2021cuad} & 31.2GB/10.4GB & 3.96B/1.31B & 488K/163K \\
         \hline
         U.S. Congressional Hearings & 6.17GB/2.06GB & 761.12M/250.04M & 24K/8K  \\
         \hline
         U.S. Tax Court PLR Corpus~\citep{taxcorpus} & 639.03MB/213.01MB & 84.25M/27.62M & 41K/14K \\
         \hline
         European Parliament Proceedings Parallel Corpus~\citep{koehn2005europarl} & 302.71MB/100.90MB & 41.55M/13.70M & 7K/2K \\
         \hline
         U.N. General Debate Corpus~\citep{baturo2017understanding} & 134.90MB/44.97MB & 17.68M/5.81M & 6K/2K  \\
         \hline
         Reddit r/legaladvice \& r/legaladviceofftopic & 299.04MB/99.68MB & 40.42M/13.56M & 110K/37K  \\
         \hline
         Bar Exam Outlines & 1.18MB/0.39MB & 123K/43K & 44/15 \\
         \hline
         Open Source Casebooks & 87.09MB/29.03MB & 9.20M/3.91M & 52/14 \\
         \hline
         Total & $\sim$ 256GB & $\sim$ 30B & $\sim$ 10M \\
         \bottomrule
    \end{tabular}}
    
    \label{tab:docsizes}
\end{table}

\newpage

\begin{longtable}{p{7cm}p{8cm}}
\caption{Filtering Norms by Data Source in the Pile of Law}\\
    \toprule
         Data Source & Examples of Filtering Norms \\
         \midrule
         Court Listener Opinions & FRCP 49.1 (requiring partial redactions for social-security number and taxpayer-identification number, date of birth, minor's names, financial account numbers; governing sealing and redaction standards for other information that parties may wish to keep private); State Rules for filing pseudonymously\footnote{\url{https://withoutmyconsent.org/50state/filing-pseudonymously/federal/}}. Judicial code of ethics govern conduct of judges; American Bar Association Model Rules of Professional Conduct govern attorney conduct. \\
         \hline
         Court Listener Docket Entries and Court Filings & \emph{Id.} \\
         \hline
         U.S. Supreme Docket Entries and Court Filings & \emph{Id.} \\
         \hline
         U.S. Board of Veterans' Appeals Decisions & 38 CFR 20.1301(c) (``Appeals on or after January 1, 1992, are electronically available for public inspection and copying on the Board's website. All personal identifiers are redacted from the decisions prior to publication.'') \\
         \hline
         U.S. Federal Trade Commission Advisory Opinions & 16 CFR 1.4 (``Written advice rendered pursuant to this section and requests therefor, including names and details, will be placed in the Commission's public record immediately after the requesting party has received the advice, subject to any limitations on public disclosure arising from statutory restrictions, the Commission's rules, and the public interest. A request for confidential treatment of information submitted in connection with the questions should be made separately.'')  \\
         \hline
         U.S. National Labor Relations Board Decisions & The U.S. National Labor Relations Board (NLRB) protects information in accordance with the Privacy Act of 1974, the E-Government Act of 2002, P.L. 107-347, and the Federal Records Act, 44 U.S.C. § 3301 et seq. Section 208 of the E-Government Act of 2002 requires all federal agencies to conduct a privacy impact assessment (PIA) for all new or substantially changed technology that collects, maintains, or disseminates personally identifiable information (PII). The goals of a PIA are to ensure conformance with applicable legal, regulatory, and policy requirements for privacy, determine privacy risks, and evaluate processes to mitigate potential privacy risks. \\
         \hline
         U.S. Department of Justice Executive Office for Immigration Review \textit{Immigration \& Nationality Decisions} & 8 CFR 208.6 (describing restrictions on disclosure of information to third parties in relation to asylum claims); 8 CFR 103.10(d) (the Attorney General may select cases to publish as precedential decisions) \\
         \hline
         U.S. Department of Labor Employees' Compensation Appeals Board & 20 CFR 501.8(c) (decisions shall be publicly available); Agency Policy not governed by law\footnote{\url{https://www.dol.gov/agencies/ecab/decisions-info}} (``To limit a claimant's exposure on the Internet, the Department of Labor (DOL) avoids referring directly to the claimant's name in decisions posted on the DOL web site on or after August 1, 2006. . . This policy is intended to protect FECA claimants from unnecessary publicity; it is not based upon a legal requirement. Neither FOIA, nor the Privacy Act, nor any other law compels DOL to take this action. Rather, this policy is based on a desire to address in a responsible way concerns raised by some claimants about the ease of access to their case decisions on the Internet. The policy seeks to comply with legal requirements to make agency decisions available on the Internet, but to do so in a way that limits a claimant's exposure to Internet users.'') \\
         \hline
         European Court of Human Rights Opinions~\citep{poudyal2020echr} & Although EHCR does not publish a formal set of PII practices, it is bound by its own interpretations of the European Convention on Human Rights, which includes among others the right to be forgotten (see Hurbain v. Belgium).\footnote{\url{https://hudoc.echr.coe.int/fre?i=001-210884} (interpreting Articles 8 and 10 of the Convention to permit the censorship of documents that infringe on the right to be forgotten).} Ordinarily, the Court appeals to publish the names of plaintiffs but to anonymize all other details.  \href{https://www.echr.coe.int/Documents/PD_anonymity_ENG.pdf}{Rules 33 and 47 of the Rules of Court} also allow for anonymity and takedown requests to be considered in the case of private data. There are no professional standards governing lawyers before the EHCR except for those imposed by the lawyer's home country \citep{sarvarian2012common}.
         \\
         \hline
         Canadian Court Opinions (ON, BC) & The names of litigants and all other case details are ordinarily public in Canadian court decisions, although in certain criminal cases pertaining to sexual offenses they may be sealed by the presiding judge (\href{https://www.ontariocourts.ca/decisions/2021/2021ONCA0013.htm}{example}). Under the Youth Criminal Justice Act, the records of criminal cases against juveniles are generally confidential.\footnote{\url{https://www.justice.gc.ca/eng/cj-jp/yj-jj/tools-outils/sheets-feuillets/recor-dossi.html}} Sealing standards in Canada, as in the U.S., permit restrictions on public access where necessary to protect important public interests \citep{sccSherman}. Canadian lawyers and judges are bound by ethical norms that prohibit discourteous speech, which impliedly includes derogatory speech \citep{lsoHandbook, bclsHandbook}.\\
         \hline
         U.S. Office of Legal Counsel Memos & Because the Office of Legal Counsel handles legal questions raised by government agencies about policy decisions, its opinions rarely pass on the particular details of a specific person's interaction with the government.\footnote{There are exceptions, see e.g. \citep{reecer2020ethical} (referring to a whistleblower complaint that was referred to OLC) but these are generally not made public.} 
         \\
         \hline
         U.S. Office of Inspector General Reports & OIG offices are effectively the ethics investigators that examine the conduct of U.S. agencies (including compliance with privacy regulations), so it would be extremely unusual for OIG offices to undertake actions that would create privacy or toxocity effects. 5 CFR \S 2638.106. \\
         \hline
         U.S. Code of Federal Regulations & Like other prospective rules of general application, the CFR does not ordinarily address individuals and thus is highly unlikely to contain private information. It is possible that the in the extremely unlikely event that PII was revealed in the CFR or Federal Register, the Privacy Act of 1974 might apply to provide a remedy. A very small number of regulations may contain archaic terms now considered toxic (see e.g. 10 C.F.R. \S § 800.003); otherwise no racial epithets are used.  \\
         \hline
         U.S. State Codes & Codified statutes (i.e. laws passed by a legislature) generally do not contain private information since they announce prospective rules that do not apply to any individual. However, state codes may include archaic laws that feature speech that would currently be classified as toxic (e.g., Miss. Code Ann. § 37-113-31). \\
         \hline
         U.S. Code & Codified statutes (i.e. laws passed by a legislature) generally do not contain private information since they announce prospective rules that do not apply to any individual. Further, most archaic references to minority communities have now been expurgated and replaced (see e.g. H.R. 4238). \\
         \hline
         U.S. Federal Rules of Evidence & Codified rules generally do not contain private information since they announce prospective rules that do not apply to any individual. The FRE contains no overt epithets. \\
         \hline
         U.S. Federal Rules of Civil Procedure & Codified rules generally do not contain private information since they announce prospective rules that do not apply to any individual. The FRCP contains no overt epithets. \\
         \hline
         U.S. Bills & As is true of the U.S. Code, proposed Bills generally do not contain private information, except in the rare cases where a bill is passed for the benefit of one individual (e.g., to \href{https://www.congress.gov/107/plaws/pvtl4/PLAW-107pvtl4.htm}{grant} a person citizenship); in such cases, a person has no expectation of privacy. Proposed bills from past eras may contain archaic derogatory language but recent legislation, which is edited and produced by professionalized offices, are unlikely to do so (\href{https://legcounsel.house.gov/about}{House}; \href{https://www.slc.senate.gov/Policies/policies.htm}{Senate}).  \\
         \hline
         U.S. Federal Register & The Federal Register contains only official communications about prospective or final rules, as well as comments submitted about those rules, other official agency action, and official actions by the President. The names and contact information that appear most frequently are individuals designated to receive public comments, whose information is personally identifiable but not sensitive in virtue of the offices they occupy. Certain executive orders may also name individuals (e.g. pardons for criminal cases). Agencies are extremely unlikely to reveal personal information outside of these contexts, though in such a highly unusual context the Privacy Act of 1974 might apply to provide a remedy. 
         \\
         \hline
         U.S. Founders Letters & Although many letters tended to use pseudonyms~\citep{shalev2003ancient}, no formal rules applied to the inclusion of information in these documents.\\
         \hline
         World Constitutions~\citep{elkins2014constitute} & No filtering standards, but many constitutions are heavily influenced by the Universal Declaration of Human Rights~\citep{law2012declining} and recent constitutions are unlikely to have offensive or private content. \\
         \hline
         EUR-Lex~\citep{chalkidis-etal-2019-large} & It is unclear if there are any standards governing European Law. However, as with U.S. legal sources it is unlikely that modern legal text has any toxic or private information since these sources promulgate laws. That being said older laws may have direct epithets and there may be indirectly toxic content in the laws.\\
         \hline
         Credit Card Agreements & Consumer Financial Protection Bureau Privacy Policy\footnote{\url{https://www.consumerfinance.gov/privacy/}}; 12 CFR 1070.13(d) (private information should be redacted by CFPB)\\
         \hline
         Terms of Service~\citep{lippi2019claudette,ruggeri2021detecting} & Since Terms of Service are not personalized, there would be no reason for PII to appear in this content. Professional norms and rules govern the drafters (attorneys) of ToS agreements (see ABA Model Rules of Professional Conduct). Reputational harms and anti-discrimination laws may also constrain overtly toxic content.\\
         \hline
         Edgar Contracts~\citep{borchmann-etal-2020-contract} & SEC Policy\footnote{\url{https://www.sec.gov/os/webmaster-faq\#reuse2}} (``My name appears in an old enforcement order or release. Is it possible to remove the document so web searches on my name don’t return the sec.gov document at the top of the results list? We don’t remove historical enforcement materials at public request or attempt to influence search result rankings. Enforcement documents from the beginning of sec.gov in 1995 remain available.'') \\
         \hline
         Atticus Contracts~\citep{hendrycks2021cuad} &  Professional norms and rules of professional conduct would govern the drafters of contracts. However, dated, offensive, and discriminatory clauses still remain in some contracts,\footnote{\href{https://www.mercurynews.com/2019/02/26/for-whites-only-shocking-language-found-in-property-docs-throughout-bay-area/}{https://www.mercurynews.com/2019/02/26/for-whites-only-shocking-language-found-in-property-docs-throughout-bay-area/}}  though they don't seem to appear in this data.\\
         \hline
         U.S. Congressional Hearings & House of Representatives 117th Congress Rule VII, clause 3 (b)(2) (``An investigative record that contains personal data relating to a specific living person (the disclosure of which would be an unwarranted invasion of personal privacy), an administrative record relating to personnel, or a record relating to a hearing that was closed under clause 2(g)(2) of rule XI shall be made available if it has been in existence for 50 years.''); Rule X, clause 11 (f) (``The select committee shall formulate and carry out such rules and procedures as it considers necessary to prevent the disclosure, without the consent of each person concerned, of information in the possession of the select committee that unduly infringes on the privacy or that violates the constitutional rights of such person. Nothing herein shall be construed to prevent the select committee from publicly disclosing classified information in a case in which it determines that national interest in the disclosure of classified information clearly outweighs any infringement on the privacy of a person.''); Rule XVII, clause 4 and 8 (unparliamentary words may be preserved for the record and may be removed only by permission or order of the house)\footnote{\url{https://sgp.fas.org/crs/misc/R45866.pdf}}\\
         \hline
         U.S. Tax Court PLR Corpus~\citep{taxcorpus} & Tax Court’s Rules of Practice and Procedure Rule 27 and 103 (governing privacy redactions and sealing); Internal Revenue Code Section 7461(b) (court can take action ``which is necessary to prevent the disclosure of trade secrets or other confidential information, including [placing items] under seal to be opened only as directed by the court.''). Despite these rules, hearings in the past have had offensive or toxic content appear.\\
         \hline
         European Parliament Proceedings Parallel Corpus~\citep{koehn2005europarl} &  \href{https://www.europarl.europa.eu/doceo/document/RULES-9-2019-07-02_EN.pdf}{Rules of Parliament Title IX Rule 226.13} (``the petitioner, a co-petitioner or a supporter may request
that his, her or its name be withheld in order to protect his, her or its privacy, in which case
Parliament shall comply with the request.'') \\
         \hline
         U.S. Supreme Court Oral Argument Transcripts &  Supreme Court Justices are not bound by a code of conduct\footnote{\url{https://sgp.fas.org/crs/misc/LSB10255.pdf}}, but professional norms likely restrict their speech.\\
         \hline
         U.N. General Debate Corpus~\citep{baturo2017understanding} & The corpus files were cleaned to remove typos and OCR conversion errors, but were not otherwise altered. But similar to other corpora of debates in political forums, the speech of U.N. General Debate speakers is likely restricted by professional norms. \\
         \hline
         Reddit r/legaladvice \& r/legaladviceofftopic & Content Moderation Policies r/legaladvice\footnote{\url{https://www.reddit.com/r/legaladvice/wiki/index\#wiki_general_rules}} (no identifying information, no illegal advice, etc.); Content Moderation Policies r/legaladviceofftopic\footnote{\url{https://www.reddit.com/r/legaladviceofftopic/}} (``personal attacks and harassing comments will be removed''; ``While it is okay to post published situations, disclosing the names or information of otherwise-anonymous parties, users, etc., is strictly forbidden.'') \\
         \hline
         Bar Exam Outlines & No restrictions are made on third party content, but given that attorneys wrote the outlines they would be governed by the same professional rules as any attorney.\\
         \hline
         Open Source Casebooks & No restrictions are made on third party content, but given that attorneys wrote the outlines they would be governed by the same professional rules as any attorney (especially because case books are meant to be educational content).\\
         \bottomrule
    
    \label{tab:licensing}
\end{longtable}

\begin{table}[H]
    \centering
        \resizebox{\textwidth}{!}{
        \begin{tabular}{p{7cm}p{7cm}}
    \toprule
         Data Source & License \\
         \midrule
         Court Listener Opinions & Underlying content is Public Domain.\\
         \hline
         Court Listener Docket Entries and Court Filings & Public Domain. \\
         \hline
         U.S. Supreme Docket Entries and Court Filings &  Public Domain. \\
         \hline
         U.S. Board of Veterans' Appeals Decisions &  Public Domain. \\
         \hline
         U.S. Federal Trade Commission Advisory Opinions &  Public Domain. \\
         \hline
         U.S. National Labor Relations Board Decisions & Public Domain. \\
         \hline
         U.S. Department of Justice Executive Office for Immigration Review \textit{Immigration \& Nationality Decisions} & Public Domain.\\
         \hline
         U.S. Department of Labor Employees' Compensation Appeals Board &Public Domain. \\
         \hline
         European Court of Human Rights Opinions~\citep{poudyal2020echr} & \href{https://www.echr.coe.int/Pages/home.aspx?p=disclaimer}{Non-commercial OK. Commercial use requires written permission.}\\
         \hline
         Canadian Court Opinions (ON, BC) & All reproduction OK (\href{https://www.ontariocourts.ca/coa/website-policies/}{ON}, \href{https://www.bccourts.ca/Privacy\%20Statement.html}{BC}). We acknowledge that these documents are sourced from the Court of Appeal for Ontario and the Court of Appeal for British Columbia, respectively. Note that these are not official versions. \\
         \hline
         U.S. Office of Legal Counsel Memos  & Public Domain.\\
         \hline
         U.S. Office of Inspector General Reports & Underlying content is Public Domain. Complication is \href{https://github.com/unitedstates/inspectors-general}{CC0 1.0 Universal public domain dedication.}\\
         \hline
         U.S. Code of Federal Regulations & Public Domain. \\
         \hline
         U.S. Supreme Court Oral Argument Transcripts &  Public Domain.\\
         \hline
         U.S. State Codes & Public Domain.\\
         \hline
         U.S. Code & Public Domain.\\
         \hline
         U.S. Federal Rules of Evidence & Public Domain. \\
         \hline
         U.S. Federal Rules of Civil Procedure & Public Domain.\\
         \hline
         U.S. Bills & Public Domain.\\
         \hline
         U.S. Federal Register & Public Domain.\\
         \hline
         U.S. Founders Letters & \href{https://www.archives.gov/global-pages/privacy.html\#copyright}{Creative Commons CC0 1.0 Universal license} \\
         \hline
         World Constitutions~\citep{elkins2014constitute} & \href{https://www.constituteproject.org/content/about?lang=en}{CC BY-NC 3.0}\\
         \hline
         EUR-Lex~\citep{chalkidis-etal-2019-large} & \href{https://huggingface.co/datasets/eurlex}{Creative Commons Attribution 4.0 International} \\
         \hline
         Credit Card Agreements & Provide by Consumer Financial Protection Bureau in the Public Domain, but original copyright and license is unknown.  We assume publication is governed by fair use standards.\\
         \hline
         Terms of Service~\citep{lippi2019claudette,ruggeri2021detecting} &  Publicly Available, but unknown license. We assume publication is governed by fair use standards. \\
         \hline
         Edgar Contracts~\citep{borchmann-etal-2020-contract} & Unknown license. We assume publication is governed by fair use standards. \\
         \hline
         Atticus Contracts~\citep{hendrycks2021cuad} & \href{https://www.atticusprojectai.org/}{CC BY 4.0} \\
         \hline
         U.S. Congressional Hearings & Public Domain \\
         \hline
         U.S. Tax Court PLR Corpus~\citep{taxcorpus} & Underlying Content is Public Domain, Complication License is \href{https://archive.data.jhu.edu/dataset.xhtml?persistentId=doi:10.7281/T1/N1X6I4}{CC BY-NC 4.0} \\
         \hline
         European Parliament Proceedings Parallel Corpus~\citep{koehn2005europarl} & \href{https://www.statmt.org/europarl/}{No copyright restrictions on compliation.} \href{https://joint-research-centre.ec.europa.eu/language-technology-resources/dcep-digital-corpus-european-parliament_en}{Non-commercial for underlying data.} \\
                  \hline
         U.N. General Debate Corpus~\citep{baturo2017understanding} & \href{https://dataverse.harvard.edu/dataset.xhtml?persistentId=doi:10.7910/DVN/0TJX8Y}{Public Domain} \\
         \hline
         Reddit r/legaladvice \& r/legaladviceofftopic &  \href{https://zenodo.org/record/3608135}{Creative Commons Attribution 4.0 International}. \href{https://www.reddit.com/wiki/api-terms}{Reddit also grants a license to copy and display the underlying data.}  \\
         \hline
         Bar Exam Outlines & Publicly Available, but unknown license. We assume publication is governed by fair use standards.\\
         \hline
         Open Source Casebooks &  All CC, varying on exact restrictiveness. Most restrictive: CC BY-NC-SA 4.0. All licensing information preserved in individual documents. \\
         \bottomrule
    \end{tabular}}
    \caption{Content licenses with links where applicable. All government-generated content in the United States is public domain.}
    \label{tab:my_label}
\end{table}

\subsection{Distribution of Topics}

\begin{figure}[H]
    \centering
    \includegraphics[width=.9\textwidth]{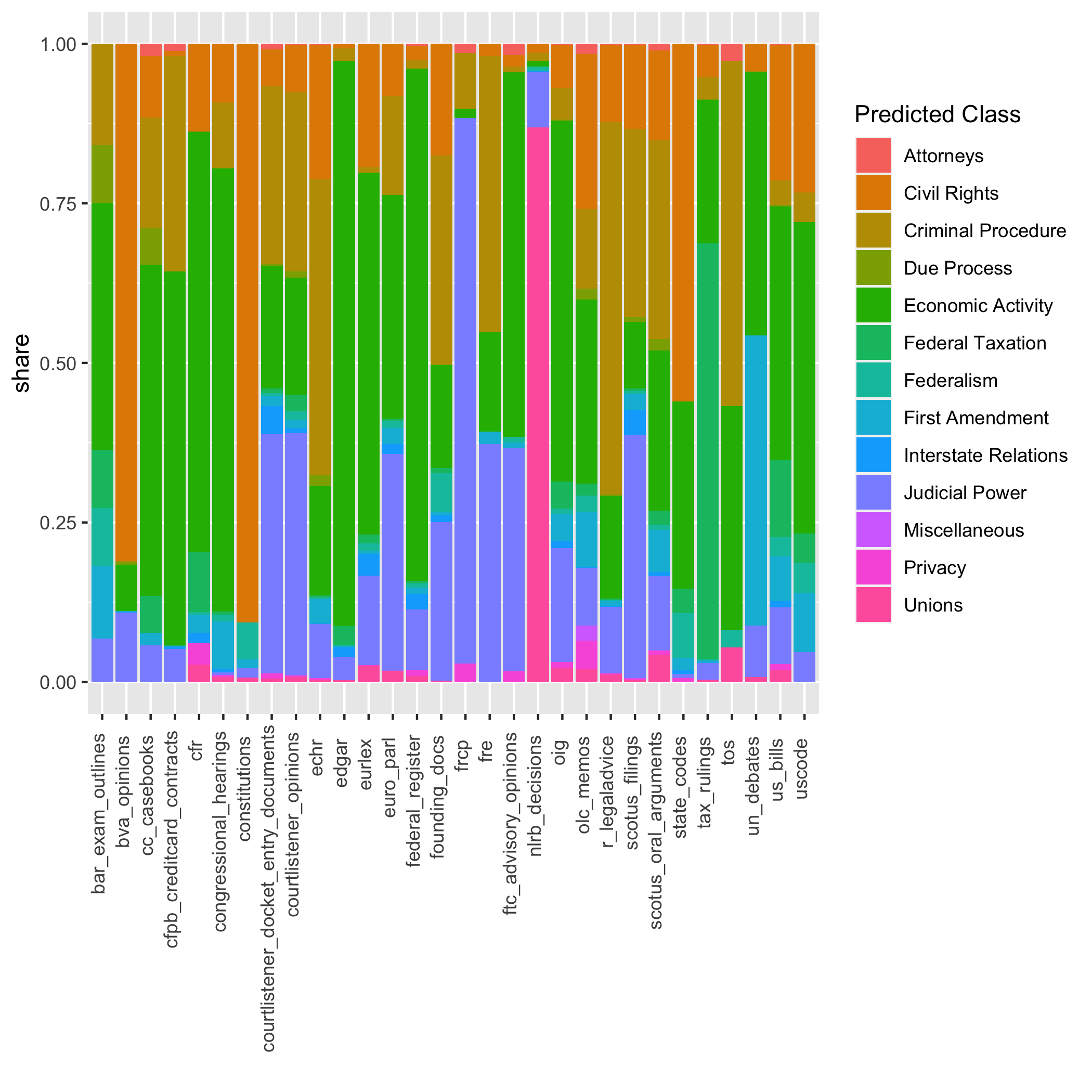}
    \caption{The distribution of topics in Pile of Law.}
    \label{fig:topics}
\end{figure}

We train a TF-IDF SVM on the LexGlue Supreme Court topic modeling task~\citep{chalkidis2021lexglue,spaeth2013supreme}. The task predicts 13 topics from case data. We then run this on our data to get a sense for topic distribution. This simple method aligns with expectations, as seen in Fgure~\ref{fig:topics}. NLRB opinions, for example, are mainly classified as ``Unions'' (as one would expect), constitutions are mainly concerned with civil rights, the federal rules of civil procedure are mostly related to judicial power, and SEC Edgar filings are mostly related to economic activity. However, there are some unintuitive categorizations that can be explained by the coding scheme of \citet{spaeth2013supreme}. 

\section{Models}
\label{app:models}

We use the \href{https://sambanova.ai/products/dataflow-as-a-service/}{SambaNova Systems Dataflow-as-a-Service\textsuperscript{TM}} platform for training an initial experimental BERT-large baseline model from scratch. The API interface abstracts away training details for training transformer models and uses SambaNova RDU chips which are optimized for data-intensive workflows, like those required for training a BERT-large model.
We implement a customized pre-processing method similar to Roberta pretraining (described below).
Despite using a training method consistent with prior BERT-like model training protocols and similarly consistent pre-processing, we found training a model on such varied data surprisingly challenging, and detail some discussion below that may be a path forward for interesting pretraining research on Pile of Law.

First, we fit a custom word-piece vocabulary to the training split of Pile of Law consisting of 29k tokens using the HuggingFace WordPiece tokenizer.\footnote{\url{https://github.com/huggingface/tokenizers}} We supplement this with a random set of legal terms from Black's Law Dictionary to make a total vocabulary size of 32k tokens.

We do not use the NSP task of BERT~\citep{devlin2018bert}, but rather use the method of Roberta~\citep{liu2019roberta}.
We train on 512 length sequences for the entirety of training and use the 80-10-10 masking, corruption, leave split of BERT~\citep{devlin2018bert}. We use a replication rate of 20 to create different masks for each context.
To generate sequences we use the LexNLP sentence segmenter (which handles legal citations which are often falsely mistaken for sentences)~\citep{bommarito2021lexnlp}. We then fill sentences until they comprise 256 tokens. We add a [SEP] token after this, and then fill sentences such that the entire span is under 512 tokens. If the next sentence in the series is too large we do not add it and fill the context with padding tokens.

For pretraining, we create a smaller training set use a randomly generated $\sim$30GB sample of the Pile of Law sampled evenly across subsets of data. This means that some subsets of data were included in their entirety while others only included a small portion of the total training data.

At first, we randomly shuffled data, but found that training this way was quite unstable and we were unable to get any model to converge to a reasonable optimum. Instead, we had to distribute data sources evenly across shards such that each device saw the same mixture of data. 

Even with this, we found that a large batch size and learning rate created large instabilities, potentially due to the diversity of the data. As a result, we used a very small learning rate (5e-6) and batch size (128) which yielded stable training. We ran this for roughly two weeks until 1.7M timesteps. Since we had leftover compute due to the small batch size, we ran two parallel model training runs with different random seeds and select the lowest log likelihood model for fine-tuning. Since we use such a low learning rate and batch size to keep training stable, given limited compute availability, we believe the model may be undertrained.

While we make both models freely available at \url{https://huggingface.co/pile-of-law/legalbert-large-1.7M-1/} and \url{https://huggingface.co/pile-of-law/legalbert-large-1.7M-2/}, we only select the lowest perplexity model for fine-tuning and call this PoL-BERT-Large. We will make intermediate checkpoints, taken at 50k step intervals, available on request.
Model Cards~\citep{mitchell2019model} are hosted on the HuggingFace website along with the models.
We evaluate the checkpoint on the CaseHOLD legal reasoning task~\citep{zheng2021does} and use the train/validation/test split from \citep{chalkidis2021lexglue}. Table~\ref{tab:casehold} shows the results. We run fine-tuning using a small hyperparameter search for those marked as (tuned). All other results are those reported by \citep{chalkidis2021lexglue}.

\begin{table}[]
    \centering
    \begin{tabular}{c|c}
    \toprule
        Model & CaseHOLD (F1)\\
        \midrule
        CaseLaw-BERT (tuned) & 78.5\\
        CaseLaw-BERT~\citep{chalkidis2021lexglue,zheng2021does} & 75.4\\
        PoL-BERT-Large (tuned) & 75.0\\
        Bert-Large-Uncased (tuned) & 71.3 \\
        \bottomrule
    \end{tabular}
    \caption{PoL-BERT-Large compared against results on the CaseHOLD \citep{zheng2021does} variant provided by \citep{chalkidis2021lexglue}.}
    \label{tab:casehold}
\end{table}

The Pile of Law model does not significantly outperform a BERT-base model trained exclusively on case law data and using a highly in-domain vocabulary for the CaseHOLD task. This is consistent with a number of recent results that suggest that masked-language model pre-training efficacy may saturate~\citep{abnar2021exploring}, especially as more diverse or distinct data sources are added~\citep{pfeiffer2022xmod}, or may only give significant gains for highly in-domain data~\citep{zheng2021does}. Because Pile of Law has an extremely diverse set of data, it may require more complicated techniques than MLM to boost performance. Moreover, due to the low learning rate we had to use, it may be possible that training for longer is essential for improved gains. We were unable to complete a full epoch over the data.
Nonetheless, we see roughly the same performance as reported in \citep{chalkidis2021lexglue} and improve over a Bert Large model.

We suspect that these results mean that the choice of vocabulary, pre-train time, and data selection procedure may all play an important role in adaptation. We hope that the BERT-large model provides an initial baseline for Pile of Law pretraining.
\section{Copyright}
\label{app:copyright}
Briefly, copyright is of particular concern as a legal risk for pretraining models \citep{bommasani2021foundation}. While most data in Pile of Law is public domain or under permissive licenses, copyrighted material may appear in the data. While most subsets of the data are extremely unlikely to contain copyrighted data, one source which may contain such material are CourtListener docket entries. In this case attorneys may have submitted exhibits to the court containing such material. For example in \emph{Dr. Seuss Enters., L.P. v. ComicMix LLC},
983 F.3d 443 (9th Cir. 2020), entire books are available as exhibits (Ex. 4 and Ex. 5). While the use and release of this data as part of court proceedings and legal investigations is protected under fair use in the United States~\citep{lemley2020fair}, if the data is then used to generate books or other content that competes with the copyright material it may create some risk of infringement. Or if used in a country without a fair use doctrine, there may be infringement concerns. As such, we suggest that the docket entries subset be omitted for training in these particular scenarios.

Finally, while  r/legaladvice, r/legaladviceofftopic, Edgar contracts, and Credit Card Agreements data were published by third parties under a CC license, it may be possible that the underlying data is under a different license. Similarly, bar exam outlines and Terms of Service contracts are under an unknown license, but were made generously publicly available by the creators (links are in the data). For these data sources, the risk of infringement is low and fair use standards should govern in the United States. However, if users wish to be extremely cautious, they may omit these subsets of data as well.
\section{Experiment Details}

\subsection{EOIR Privacy}
\label{app:eoir}

\begin{figure}[H]
    \centering
    \includegraphics[width=\textwidth]{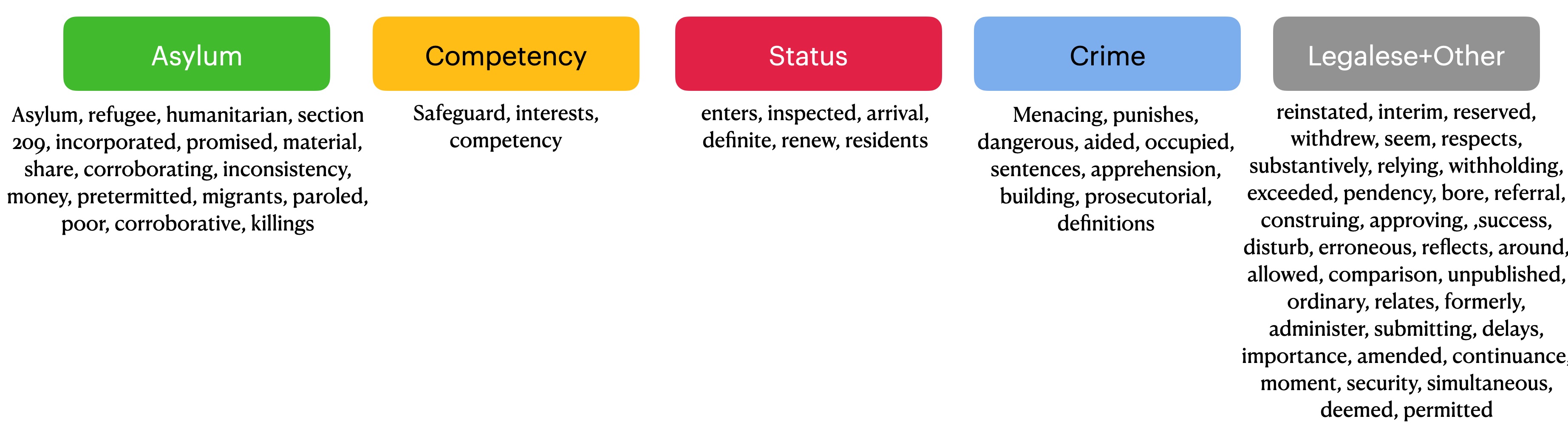}
    \caption{A causal lexicon learned for the EOIR privacy task, manually sorted by topic with contextual information. Year was used as a control variable.}
    \label{fig:vocab_full}
\end{figure}

To construct the EOIR dataset from which we train, we first split EOIR opinions into paragraphs. For each paragraph we mask any references to the respondent, including terms like ``respondent'' and ``appellant.'' We then extract using regex expressions whether the court used a pseudonym. This is extracted from the header of the file. For example, ``Matter of A-B-C-'' would be marked as using a pseudonym. From this we set the label to be either pseudonym or no pseudonym. We then train a distillbert model to take the masked paragraph as input and output a prediction of whether the text should be pseudonymized.  We provide the masked and labeled data in \url{https://huggingface.co/datasets/pile-of-law/eoir_privacy}. As well as the trained distillbert model here: \url{https://huggingface.co/pile-of-law/distilbert-base-uncased-finetuned-eoir_privacy}. Figure~\ref{fig:vocab_full} is an extended figure showing all causal terms learned from this data based on hand-crafted clustering. We describe some reasons for the cluster groupings below.

\paragraph{Reasoning for unobvious cases.} While some connections from vocabulary to topic might be obvious, we provide some reason for classification of non-obvious connections in the following Table:

\begin{table}[H]
    \centering
    \begin{tabular}{c|p{10cm}}
    \toprule
         Word & Group Reasoning \\
         \midrule
         section 209 & INA \S 209 deals with asylum claims\\
         inspected & There is a standard that immigrants be inspected and admitted or paroled on entry into the U.S.\\
         migrants & all contextualized references referred to Cuban and Haitian migrants seeking asylum\\
         killings & all references were to extrajudicial or mass killings commonly referred to in asylum claims\\
         poor & all references referred to the poor conditions of the country of origin in asylum claims\\
         pretermitted & under 8 CFR § 1208.13, Immigration Judges can pretermit an asylum application without a hearing\\
         money & all references were to threats of extortion for money, under threat of violent punishment, directed at asylum claimants by figures in their country of origin\\
         inconsistency & immigration judges look for inconsistencies in credible fear interviews when evaluating asylum claims\\
         promised & all references were to promises made to extortionary figures by asylee claims (e.g., the asylum seeker promised to join a gang if their life was spared, and subsequently left the country to seek refuge)\\
         material & ``a material element'' of the asylum application was false\\
         share & when evaluating asylum claims, one determination is whether the applicant is part of a group ``composed of members who share a common immutable characteristic.'' Matter of M-E-V-G-, 26 I\&N Dec. 227 (BIA 2014).\\
         safeguard & all references were to procedural safeguards put in place when a person's competency to stand trial was in question\\
         building,occupied & all references were to burglary\\
         interest & all references were to competing interests during competency evaluations\\
         corroborating,corroborative & asylum claims are sometimes evaluated based on corroborating/corroborative evidence for the contents of the application\\
    \end{tabular}
\end{table}

\subsection{Mean Toxicity Scores, by Supreme Court Issue Area}

Note for toxicity experiments we use the Harvard Case Law Access Project meta-data to associate opinion text to a given date~\citep{documentation12caselaw}.

\begin{figure}[H]
    \centering
    \includegraphics[width=\textwidth]{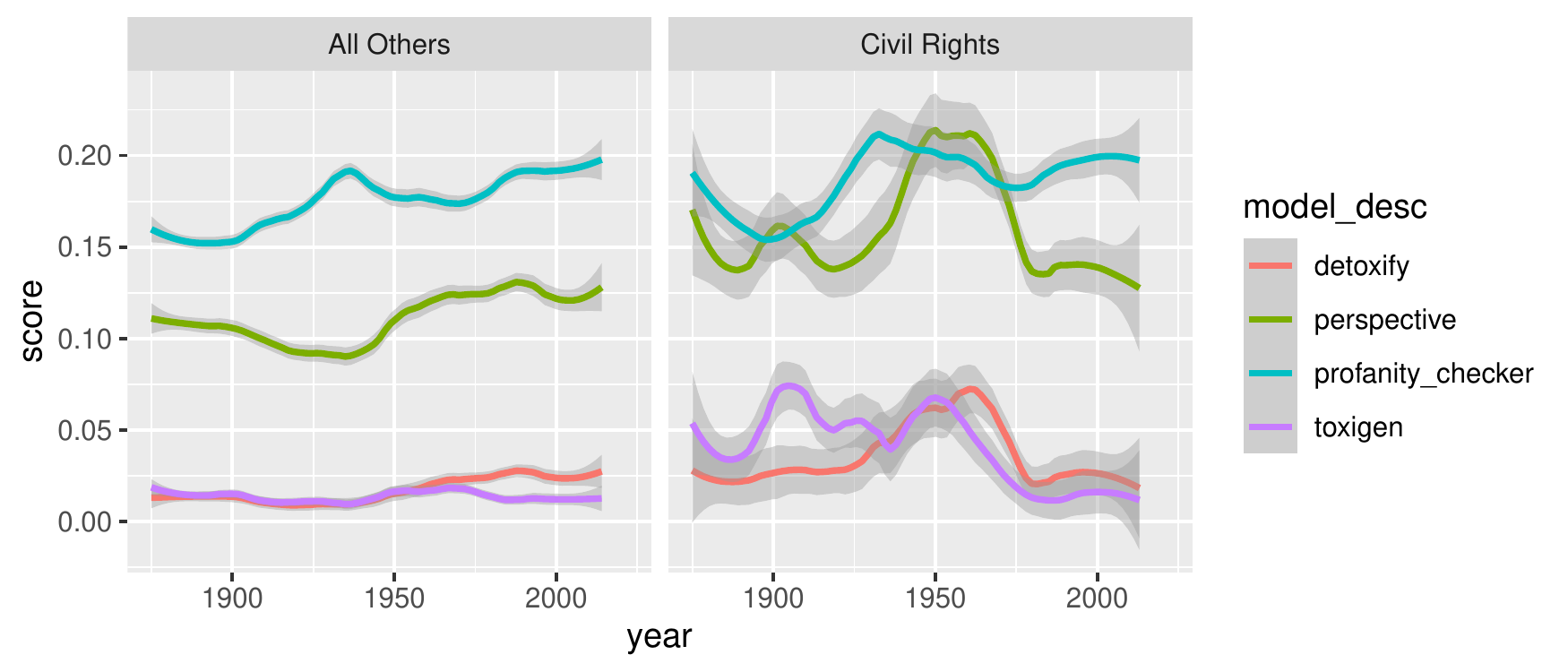}
    \caption{Mean toxicity of `highly toxic' sentence in the average Supreme Court opinion, by issue type and year, 1875-present.}
    \label{fig:scdb_issues_toxicity}
\end{figure}

In Figure~\ref{fig:scdb_issues_toxicity}, we provide more detail on the discussion in Section 4.2 regarding the vulnerability of toxicity scores to factual circumstances. For each Supreme Court opinion beginning in 1875, we obtain the sentence in the 97.5th percentile of toxicity as indicated by each of the four filters we study. These sentences represent ``highly toxic'' sentences within each opinion, while leaving out the 2.5\% most extreme cases to reduce sampling variability. We then average the toxicity scores of these highly toxic sentences by year and by whether or not they are categorized as pertaining to `civil rights' by \citep{spaeth2013supreme}. 

The right panel suggests that the onset of the Civil Rights Era in the late 1940s corresponds to a rise in the assessed toxicity of Supreme Court cases, despite the fact that many of the cases containing highly toxic sentences were instrumental in dismantling official segregation. For example, the 97.5th percentile of toxicity in \textit{Brown v. Board of Education}, which ended the official segregation of schools, is .535 as rated by the Perspective API. 

\newpage
\subsection{Cohen's Kappa for All Models}
\begin{figure}[H]
    \centering
    \includegraphics[scale=0.8]{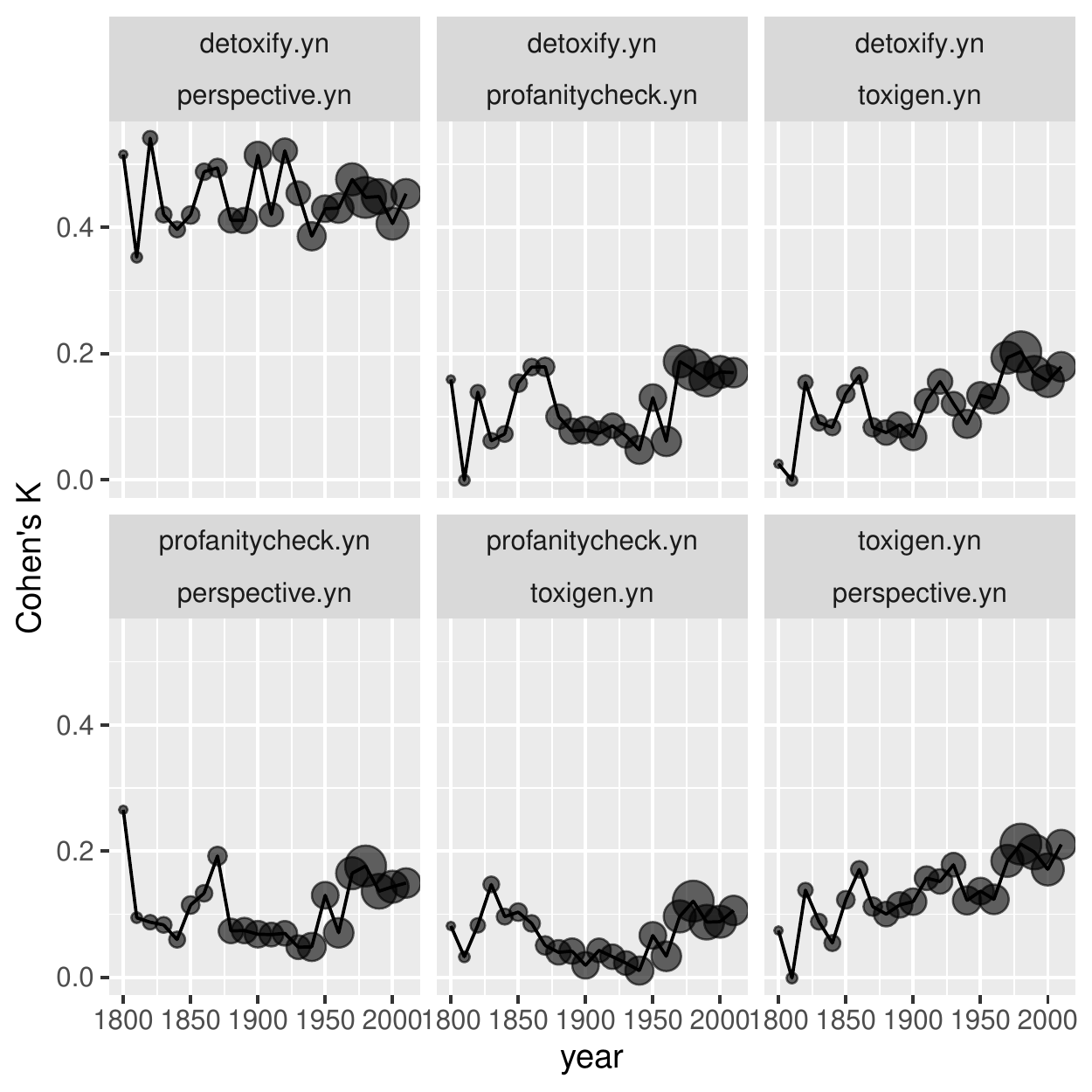}
    \caption{Cohen's $\kappa$ for all pairs of models studied. Each bin corresponds to the sentences in Supreme Court cases decided over the nearest ten years. Note: detoxify and perspective may have higher agreement since they appear to be trained on similar data sources.}
    \label{fig:cohens_complete}
\end{figure}

\newpage
\subsection{Qualitative Examples}

\begin{center}
    \textbf{{\color{red}Content Warning:} This section contains quotations that may be offensive or upsetting.}
\end{center}

\begin{table}[H]
\setlength{\tabcolsep}{4pt} 
    \renewcommand{\arraystretch}{1.1}
    \centering
    \caption{Context length perturbation. \small{Adding additional context can make deeply offensive content be labeled as non-toxic using Toxigen~\citep{hartvigsen2022toxigen}. Red indicates the toxic sentence with score paired in red. Black font indicates the appended sentence and toxicity score for the entire paragraph. We redact identifying information and epithets for presentation when applicable. Top two are hand-identified, bottom are randomly samples toxic sentences according to Toxigen.}}
    \begin{tabular}{L{0.1\linewidth}p{0.8\linewidth}p{0.1\linewidth}}
    \toprule
    \textbf{Category} & \textbf{Quote} & \textbf{Score} \\
    \midrule
      True Use & \small{``{\color{red} There is a race so different from our own that we do not permit those belonging to it to become citizens of the United States. Persons belonging to it are, with few exceptions, absolutely excluded from our country. I allude to the Chinese race.} But, by the statute in question, a Chinaman can ride in the same passenger coach with white citizens of the United States, while citizens of the black race in Louisiana, many of whom, perhaps, risked their lives for the preservation of the Union, who are entitled, by law, to participate in the political control of the State and nation, who are not excluded, by law or by reason of their race, from public stations of any kind, and who have all the legal rights that belong to white citizens, are yet declared to be criminals, liable to imprisonment, if they ride in a public coach occupied by citizens of the white race. It is scarcely just to say that a colored citizen should not object to occupying a public coach assigned to his own race.''}  & {\color{red}0.78}/0.03\\
       \hline
      True Use & \small{``{\color{red} The appellant, [NAME REDACTED], age [AGE REDACTED] years, was a homosexual p[****]t.} The evening of [DATE REDACTED], the decedent was in a tavern of a hotel in [LOCATION REDACTED]. [NAME REDACTED], who had a room in this hotel, also came into the tavern, and in a short time took a seat by the decedent and engaged him in conversation.''} & {\color{red} 0.98}/0.07\\
      \midrule
      Mention (quoted) & \small{``It was buzzards of the [NAME REDACTED] stripe who shot down my buddy. It was Barnett gunmen who staged the assault on [NAME REDACTED]. {\color{red} It is Jew thugs who have `pulled' practically every robbery in this city.} It was a member of the Barnett gang who shot down [NAME REDACTED] while he stood in the shelter of [NAME REDACTED]'s ham-cavern on [NAME REDACTED] avenue.''} & {\color{red} 0.99}/0.00\\
      \hline
      Mention (quoted)& \small{`` ` Other signs would most naturally have been understood as suggesting — falsely—that [NAME REDACTED] was gay. Homosexuality was the theme of many of the signs. There were signs reading “God Hates F[***],” “Semper Fi F[***],” “F[***] Doom Nations,” and “F[**] Troops.” Id., at 3781-3787.' ''} & {\color{red} 0.99}/0.01\\
      \hline
      True Use & \small{``And one thing that we very often say and talk about is the three classes of liars. There is the plain liar, the damn liar, and the expert witness. {\color{red}And of all of them, the expert witness is the worst.} ``There were a few of them here.''''}  & {\color{red} 0.91}/0.00\\
      \hline
      Mention (quote) & \small{`` `The two statutes under which appellants were convicted and sentenced are part of a comprehensive statutory scheme aimed at prohibiting and punishing interracial marriages. The Lovings were convicted of violating \S 20-58 of the Virginia Code: ``Leaving State to evade law. {\color{red} -- If any white person and colored person shall go out of this State, for the purpose of being married, and with the intention of returning, and be married out of it, and afterwards return to and reside in it, cohabiting as man and wife, they shall be punished as provided in § 20-59, and the marriage shall be governed by the same law as if it had been solemnized in this State.} The fact of their cohabitation here as man and wife shall be evidence of their marriage.''' ''} & {\color{red} 0.82}/0.40\\
      \hline
      Ambiguous & \small{``Respondents inexplicably make no effort to address Chapter 2 under the Agostini test. Instead, dismissing Agostini as factually distinguishable, they offer two rules that they contend should govern our determination of whether Chapter 2 has the effect of advancing religion. {\color{red}They argue first, and chiefly, that “direct, nonincidental” aid to the primary educational mission of religious schools is always impermissible.} Second, they argue that provision to religious schools of aid that is divertible to religious use is similarly impermissible.' ''} & {\color{red}0.77}/0.01\\
       \bottomrule
    \end{tabular}
    \label{tab:toxigen_context_length}
\end{table}

\section{Details on Comparative Law}

\label{app:law_comp}

\begin{table}[H]
\caption{Availability of Identifying Information Across Administrative Settings}
\begin{tabular}{>{\raggedright}p{0.15\linewidth}p{0.25\linewidth}p{0.25\linewidth}p{0.25\linewidth}}
\toprule
Jurisdiction & \textbf{Civil Cases} & \textbf{Criminal Cases} & \textbf{Juvenile Data} \\
 \midrule
\textbf{U.S. Federal Courts} & Generally available, \textit{except} dates of birth, financial account numbers, or social security numbers [Fed. R. Civ. Pr. 5.2(a)]. Cases are sealed under exceptional circumstances. & Generally available once filed and, for federal crimes, are never sealed. Filings cannot include dates of birth, SSNs, or residential addresses. [Fed. R. Crim. Pr. 9037(a).] & Juvenile criminal records are generally confidential [18 U.S.C. \S 5038].  Juvenile names must be partially redacted from civil records [Fed. R. Civ. Pr. 5.2(a)(3)].\\
& & &  \\ 
\textbf{U.S. Administrative Agencies} & Identifying information, including names, generally omitted from public records. [5 U.S.C. 552(a)]  & N/A & Disclosure not explicitly forbidden, but likely more stringent than adult disclosure. [See text] \\
& & &  \\ 
\textbf{German Courts} & Public judgments exclude all identifying information. [GVG \S 174]  &  Most criminal records are automatically expunged five years after the completion of the sentence or three years after death. & Juvenile criminal records are excluded entirely from federal data unless the conviction results in a sentence of over 1 year; even then access is restricted. \citep{matthews2018youth}\\
& & &  \\ 
\textbf{Chinese Courts} & The names of litigants and case details are public for most cases except if they fall into an excluded category, such as disputes resolved via mediation \citep{liebman2020mass}.  &  The names of litigants and case details are public for criminal cases as of 2016 \citep{liebman2020mass}.  & Juvenile criminal records are categorically exempt from disclosure \citep{liebman2020mass}. \\
& & &  \\ 
\textbf{Canadian Courts} & The names of litigants and case details are public for most cases unless they meet the standard for sealing.  &  Criminal records are generally public, but all criminal records are eligible for `suspension' (a form of sealing) after a certain period of good behavior. [Criminal Records Act]  & Youth criminal records are always confidential and are automatically sealed after the conclusion of the sentence [Youth Criminal Justice Act]. \\
& & &  \\ 
\bottomrule
\end{tabular}
\label{tab:comparative_privacy_large}
\end{table}

\section{Qualitative Examples}

We include some qualitative examples from the data to better understand its nature. For example, an OLC Memo might look like the following:

\begin{displayquote}
Constitutionality of the Qui Tam Provisions of the False Claims Act Qui tam suits brought by private parties to enforce the claims of the United States violate the Appointments Clause of the Constitution because qui tam relators are ``Officers of the United States" but are not appointed in accordance with the requirements of the Appointments Clause. Private qui tam actions violate the doctrine of Article III standing because the relator has suffered no personal ``injury in fact.'' The qui tam provisions of the False Claims Act violate the separation of powers doctrine because they impermissibly infringe on two aspects of the President's authority to execute the laws: the discretion whether to prosecute a claim and the authority to control the conduct of litigation brought to enforce the Government's interests. Given qui tam's clear conflict with constitutional principles, any argument . . . .\footnote{\url{https://www.justice.gov/sites/default/files/olc/opinions/1989/07/31/op-olc-v013-p0207_0.pdf}}
\end{displayquote}

An excerpt from the US Bills looks like:

\begin{displayquote}

113 S2875 IS: National Guard Investigations Transparency and Improvement Act of 2014

U.S. Senate

2014-09-18

Pursuant to Title 17 Section 105 of the United States Code, this file is not subject to copyright protection and is in the public domain.

II

113th CONGRESS

2d Session

S. 2875

IN THE SENATE OF THE UNITED STATES
September 18, 2014

Mr. Begich introduced the following bill; which was read twice and referred to the Committee on Armed Services
A BILL

To codify in law the establishment and duties of the Office of Complex Administrative Investigations in the National Guard Bureau, and for other purposes.

Be it enacted by the Senate and House of Representatives of the United States of America in Congress assembled,

1. Short title.

This Act may be cited as the National Guard Investigations Transparency and Improvement Act of 2014.

2. Codification in law of establishment and duties of the Office of Complex Administrative Investigations in the National Guard Bureau.

(a) In general.—There is in the Office of the Chief of the National Guard Bureau the Office of Complex Administrative Investigations (in this section referred to as the “Office”) . . . .\footnote{\url{https://www.congress.gov/bill/113th-congress/senate-bill/2875/text?r=9&s=1}}

\end{displayquote}

\end{document}